%% file: acl_latex.tex
\documentclass[11pt]{article}

\usepackage[preprint]{acl}

\usepackage{times}
\usepackage{latexsym}
\usepackage{amssymb}
\usepackage{amsmath}
\usepackage[dvipsnames]{xcolor}
\usepackage[T1]{fontenc}

\usepackage[utf8]{inputenc}

\usepackage{microtype}

\usepackage{inconsolata}

\usepackage{graphicx}
\usepackage[utf8]{inputenc}
\usepackage[ruled,vlined]{algorithm2e}
\usepackage{float}
\usepackage{booktabs}
\usepackage{multirow}
\usepackage{subcaption}
\usepackage[most]{tcolorbox}
\usepackage[version=4]{mhchem}
\usepackage[table]{xcolor}   
\title{\raisebox{-0.3\height}{
    \includegraphics[height=2.3em]{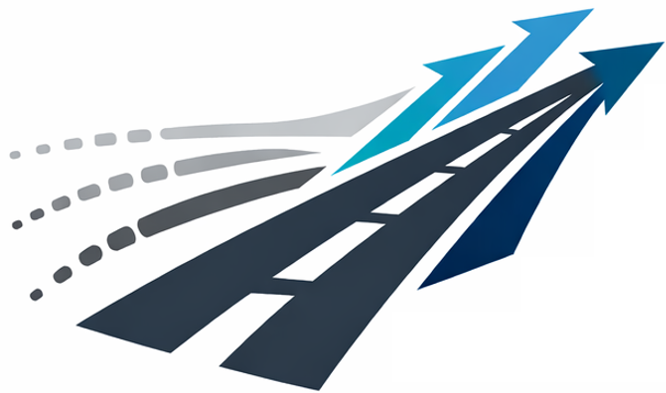}
}
\hspace{-1em}
\textcolor[RGB]{44,58,73}{ETR: } Entropy Trend Reward for Efficient Chain-of-Thought Reasoning}

\newtcolorbox{promptbox}{colback=blue!5, colframe=blue!75!black, title=Question}
\newtcolorbox{thoughtbox2}{colback=green!5, colframe=green!60!black, title=Reasoning Trajectory (1149 Tokens), breakable}
\newtcolorbox{answerbox}{colback=green!5, colframe=green!60!black, title=Final Answer}

\newtcolorbox{thoughtbox}{colback=green!5, colframe=green!60!black, title=Reasoning Trajectory (667 Tokens), breakable}
\newtcolorbox{errorbox}{
    colback=red!5,       
    colframe=red!75!black, 
    title=Info,   
    fonttitle=\bfseries,   
    breakable              
}
\newtcolorbox{thoughtbox3}{colback=green!5, colframe=green!60!black, title=Reasoning Trajectory (940 Tokens), breakable}

\author{
  Xuan Xiong$^{1}$\quad
  Huan Liu$^{2}$\thanks{Research Lead, Corresponding Author.}\quad
  Li Gu$^{3}$\quad
  Zhixiang Chi$^{1}$\quad
  Yue Qiu$^{4}$\quad
  Yuanhao Yu$^{2}$\quad
  Yang Wang$^{3}$ \\
  \\
  $^{1}$University of Toronto \quad
  $^{2}$McMaster University \quad
  $^{3}$Concordia University \quad
  $^{4}$University of Ottawa \\
}

\begin{document}
\maketitle

\begin{abstract}
Chain-of-thought (CoT) reasoning improves large language model performance on complex tasks, but often produces excessively long and inefficient reasoning traces. Existing methods shorten CoTs using length penalties or global entropy reduction, implicitly assuming that low uncertainty is desirable throughout reasoning. We show instead that reasoning efficiency is governed by the trajectory of uncertainty. CoTs with dominant downward entropy trends are substantially shorter. Motivated by this insight, we propose \textbf{E}ntropy \textbf{T}rend \textbf{R}eward (\textbf{ETR}), a trajectory-aware objective that encourages progressive uncertainty reduction while allowing limited local exploration. We integrate ETR into Group Relative Policy Optimization (GRPO) and evaluate it across multiple reasoning models and challenging benchmarks. ETR consistently achieves a superior accuracy–efficiency trade-off, improving DeepSeek-R1-Distill-7B by +9.9\% accuracy while reducing CoT length by 67\% across four benchmarks. Our code is available at \url{https://github.com/Xuan1030/ETR}.
\end{abstract}

\input{introduction}
\input{related_work}
\input{new_method}
\input{experiments}

\input{conclusion}
\bibliography{custom}

\clearpage
\appendix

\input{appendix}

\end{document}

%% file: introduction.tex
\section{Introduction}
Large language models (LLMs) increasingly rely on chain-of-thought (CoT) \cite{cot} reasoning to solve complex tasks by decomposing them into intermediate steps. By producing intermediate  reasoning steps, LLMs can improve answer accuracy, interpretability, and robustness across complex tasks \cite{survey1, survey2, survey3}.  However, CoT often comes with significant practical drawbacks. Models tend to generate unnecessarily long, repetitive, and sometimes self-contradictory reasoning sequences before reaching a conclusion \cite{nothinking,survey3}. Such ``overthinking'' significantly increases inference latency.

Existing approaches to CoT reasoning can be broadly divided into training-free and training-based methods.
Training-free methods rely on heuristic prompt design \cite{prompt1,prompt2} or early stopping criteria \cite{yang2025dynamic} to reduce reasoning length.
Training-based methods include supervised fine-tuning \cite{sft1, sft2, sft3} and reinforcement learning \cite{O1-pruner,L1}, among which reinforcement learning–based approaches have attracted growing attention due to their stronger generalization.
A representative example is length-based reward design \cite{O1-pruner,L1}, which encourages concise reasoning by penalizing long trajectories. However, such rewards are inherently content-blind, discouraging informative intermediate reasoning steps that may be important for maintaining final accuracy \cite{huang2025pear}.

\begin{figure}[!t]
\centering
\includegraphics[width=1\linewidth]{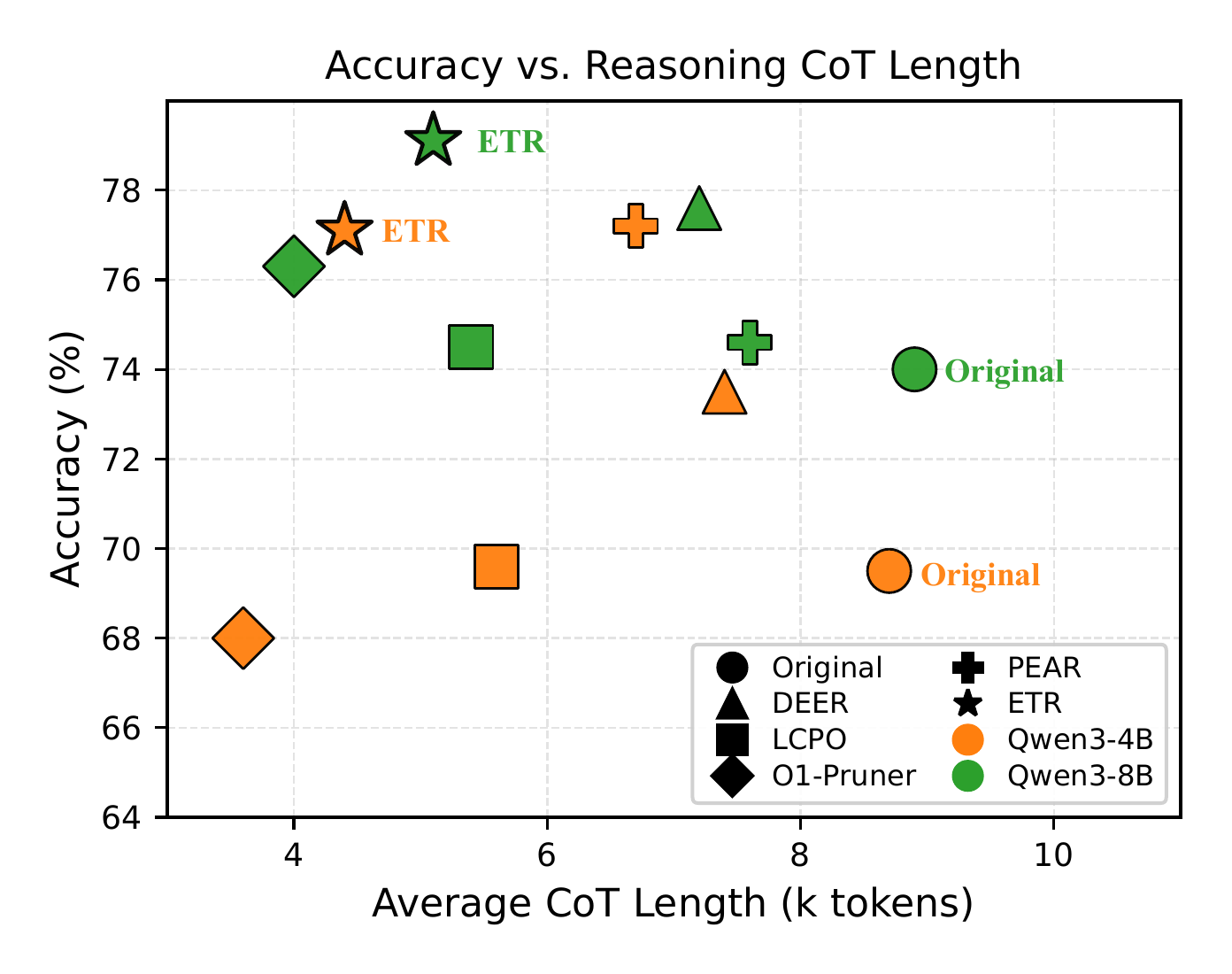}
\caption{Accuracy versus average chain-of-thought (CoT) length across model sizes. Accuracy is averaged over four representative benchmarks. Compared methods exhibit different accuracy–efficiency trade-offs, with ETR achieving strong accuracy at substantially shorter reasoning lengths.}
\vspace{-1em}
\label{fig:first}
\end{figure}

Recent work has begun to link CoT length to  predictive uncertainty, observing that CoTs with higher overall entropy tend to be longer and proposing to reduce global entropy via supervised fine-tuning \cite{step_entropy} or reinforcement learning \cite{huang2025pear}. However, this perspective implicitly assumes that low uncertainty is desirable throughout the entire reasoning process. Such uniform entropy suppression is misaligned with how effective human reasoning unfolds. For example, early steps are naturally exploratory, involving multiple plausible directions, while later stages become increasingly constrained as the solution structure emerges. Consequently, an effective CoT is not one that maintains low entropy everywhere, but one whose uncertainty progressively reduces over time, allowing occasional local backtracking while preserving a coherent global descent.
To elaborate, we provide empirical support 
for this trajectory-centric view by analyzing step-wise entropy patterns in generated CoTs. We find a clear relationship between the directionality of uncertainty evolution and reasoning length: CoTs dominated by downward entropy trends are substantially shorter, whereas those with frequent entropy increases tend to be much longer.

Motivated by this finding, we propose \textbf{E}ntropy \textbf{T}rend \textbf{R}eward (\textbf{ETR}) to optimize reasoning efficiency by explicitly shaping the trajectory of uncertainty during CoT generation. Rather than penalizing entropy uniformly, our approach encourages reasoning behaviors that achieve steady, coherent uncertainty reduction over time. This trajectory-aware objective provides dense feedback throughout the reasoning process, enabling the model to distinguish between productive exploration and inefficient uncertainty oscillation. We integrate this trajectory-aware uncertainty shaping into Group Relative Policy Optimization (GRPO) \cite{GRPO} that preserves answer correctness as a hard requirement, using efficiency signals only to compare and refine correct reasoning paths. An important consequence is an implicit, instance-adaptive stopping behavior: uncertainty collapses quickly for easy problems, yielding short CoTs, while harder problems are solved through gradual belief refinement without prolonged self-reflection or handcrafted length constraints.



In summary, this paper makes three contributions. First, we identify global uncertainty trends as a key determinant of chain-of-thought length and efficiency, shifting the focus from static entropy measures to trajectory-level reasoning dynamics. Second, we introduce a trajectory-aware uncertainty shaping approach that encourages progressive uncertainty contraction while allowing occasional local exploration. Third, we demonstrate that this approach yields shorter, more decisive reasoning traces while preserving answer quality, offering a principled path toward scalable and efficient chain-of-thought reasoning.

%% file: related_work.tex
\section{Related Work}
\subsection{Reinforcement Learning for LLM}
Reinforcement learning has emerged as a powerful paradigm for unlocking the latent reasoning capabilities of Large Language Models (LLMs). Recent breakthroughs such as DeepSeek-R1 \citep{deepseekr12025incentivizing} and OpenAI-o1 \citep{openai2024o1systemcard} demonstrate the important role of Reinforcement Learning from Verifiable Rewards (RLVR) in encouraging deep thinking and broad explorations. Despite the performance gains, the transition towards long CoT reasoning has introduced a critical challenge: the ``overthinking'' phenomenon. Recent studies \citep{survey1, survey3} reveals that models often generate excessively verbose reasoning traces, leading to a substantial inference latency and increased computational overhead. Such observation underscores the urgency of developing methods to mitigate redundancy within reasoning paths without compromising accuracy.

\subsection{Efficient Reasoning Model}

Existing approaches to efficient CoT reasoning can be broadly grouped into three categories. (1) Heuristic-guided methods design specialized prompts \cite{prompt1,prompt2} or stopping criterion \cite{yang2025dynamic} to steer models toward shorter reasoning paths without training, but their effectiveness often degrades for smaller models with limited controllability.  (2) Variable-length CoT methods \cite{sft1, sft2, sft3} rely on supervised fine-tuning with datasets containing reasoning traces of different lengths. However, they typically exhibit limited generalization beyond the training distribution. (3) length-based reward design \cite{O1-pruner, L1} incorporates explicit length rewards into reinforcement learning, encouraging concise and correct reasoning. While effective at controlling the number of generated tokens, length-based methods are fundamentally content blind. they treat all tokens equally regardless of whether they contribute useful information.

To move beyond rigid length constraints, recent work dives into entropy \citep{shannon1948mathematical} that quantifies the model's uncertainty when generating tokens.
There are training-free approaches such as CGRS \citep{huang2025efficient}, which leverage token-level entropy to suppress self-reflection tokens and mitigate response length.
Other studies \citep{agarwal2025unreasonable, huang2025pear} investigate entropy-based rewards as a signal for regulating the model's internal uncertainty during reasoning.
While existing entropy-based rewards focus on minimizing instantaneous uncertainty to streamline outputs, our entropy trend rewards monitor the evolution of entropy across the reasoning trace and introduce a more nuanced perspective.

%% file: new_method.tex
\begin{figure*}[!t]
    \centering
    \begin{subfigure}[t]{0.48\textwidth}
        \centering
        \raisebox{6pt}{
        \includegraphics[width=\linewidth]{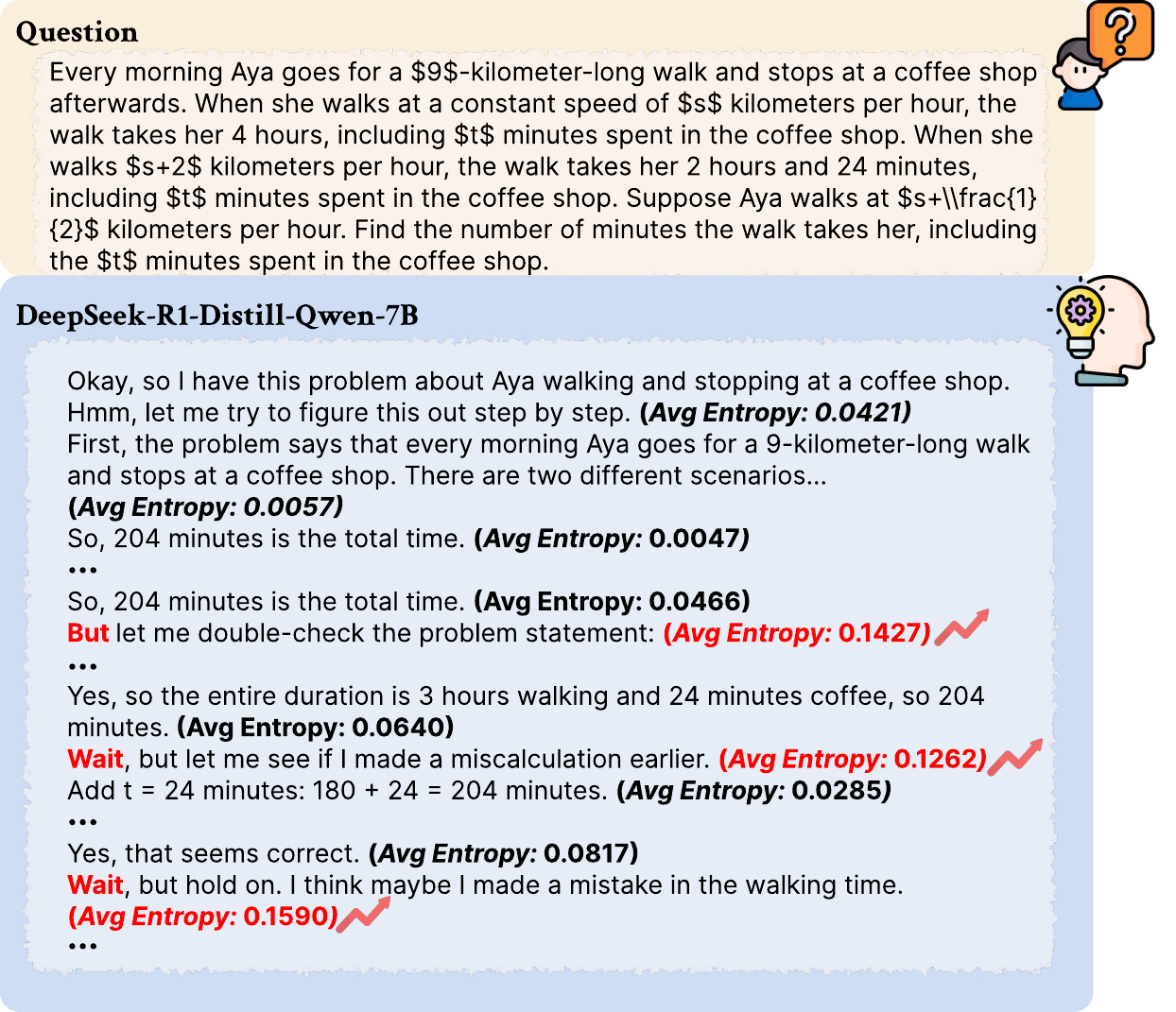}
        }
        \label{fig:left}
    \end{subfigure}
    \hfill
    \begin{subfigure}[t]{0.48\textwidth}
        \centering
        \includegraphics[width=\linewidth]{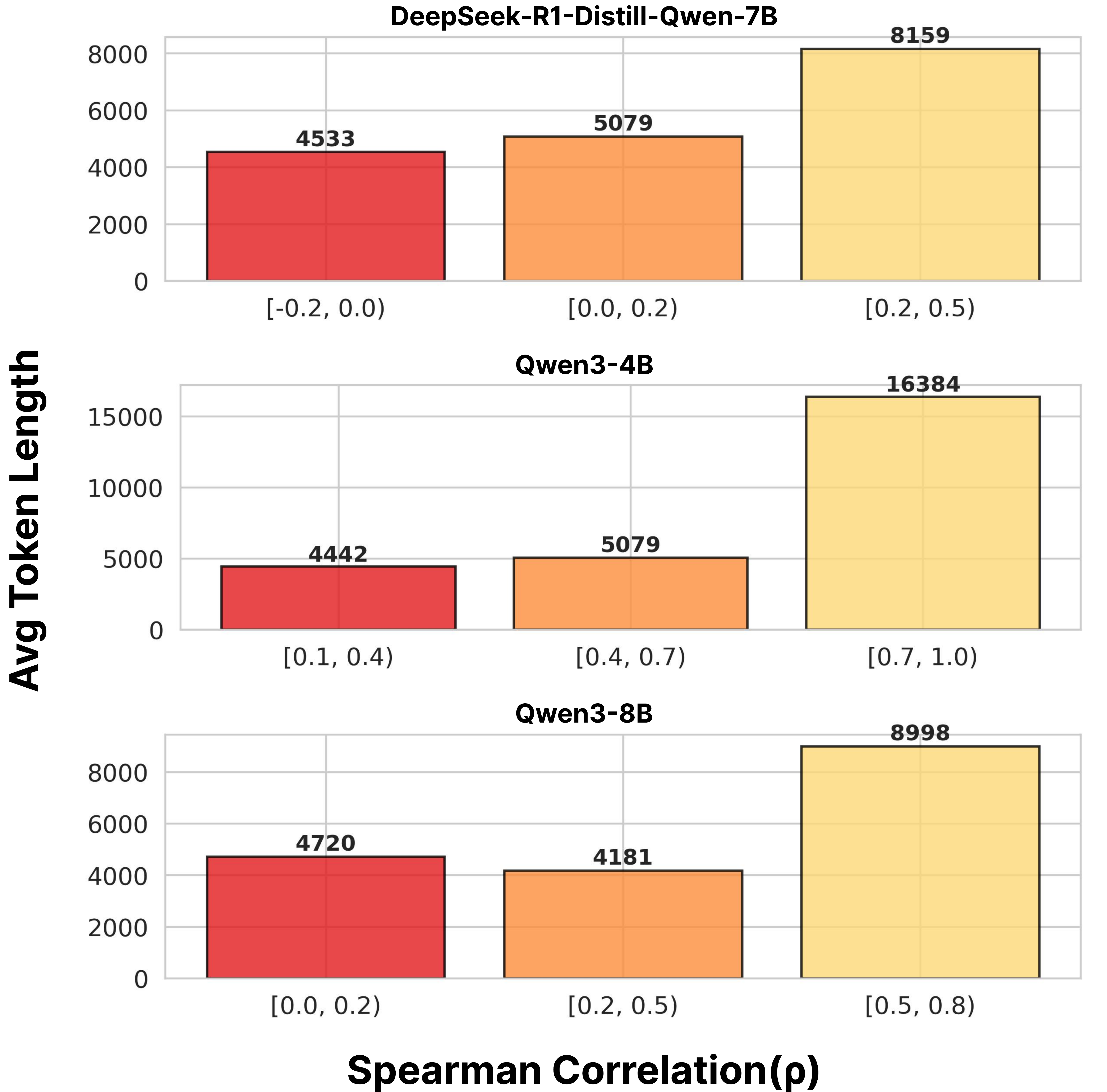}
        \label{fig:right}
    \end{subfigure}
    \vspace{-2em}
    \caption{Step-wise entropy dynamics in generated CoTs on the MATH500 dataset \cite{hendrycks2021math}. (Left) A sample response illustrating high step-wise entropy during self-reflection, marked by uncertainty cues (e.g., but, wait, etc.).
(Right) Across models, a higher Spearman rank correlation ($\rho$) between reasoning step index and step-wise entropy indicates persistent entropy growth during reasoning. This is consistently associated with longer outputs and greater token redundancy.
}    
    \label{fig:two_column}
\end{figure*}

\section{Preliminary}
We briefly review Group Relative Policy Optimization (GRPO) \citep{GRPO}, which we adopt as the policy optimization algorithm in this work.
GRPO is a policy gradient approach that evaluates model outputs
based on their relative performance within a group of sampled responses, rather than relying on explicit value function estimation \cite{ppo}.
For a given input question $q$, GRPO samples a set of $G$ candidate responses
$\{o_1, o_2, \dots, o_G\}$ from the current policy.
Each response $o_i$ is assigned a scalar return $r_i$. The return is defined as $r_i \;\triangleq\; R(q, o_i),$
where $R(q, o)$ denotes the reward assigned to response $o$ for question $q$. And a group-normalized advantage is computed as $\hat{A}_i
=
\frac{r_i - \frac{1}{G} \sum_{j=1}^{G} r_j}{\sqrt{\frac{1}{G} \sum_{j=1}^{G} (r_j - \mu_r)^2}}.$
This group-wise normalization emphasizes the relative ranking of candidate responses under the same
input and reduces sensitivity to reward scale variations across different questions.

GRPO adopts a PPO-style clipped objective for policy optimization.
Let $\pi_{\theta_{\mathrm{old}}}$ denote the policy used to generate the sampled responses, and define
the likelihood ratio
$\tau_i(\theta) = \pi_\theta(o_i \mid q) / \pi_{\theta_{\mathrm{old}}}(o_i \mid q)$.
The optimization objective is given by
\begin{equation}
\label{eq:grpo_objective}
\mathcal{J}_{\text{GRPO}}(\theta)
=
\mathbb{E}\Bigg[
\frac{1}{G}\sum_{i=1}^{G}
\min \mathcal{L}_{i,t}(\theta)
-\beta\,\mathrm{KL}(\pi_\theta)
\Bigg], 
\end{equation}
where the KL term regularizes policy updates and helps maintain stable training dynamics. $\mathcal{L}_{i,t}(\theta)$ denotes the token-level surrogate loss  given by:
\begin{equation}
    \mathcal{L}_{i,t}(\theta) = \Big(
\tau_i(\theta)\hat{A}_i,\;
\operatorname{clip}\!\Big(
\tau_i(\theta),\,
1-\epsilon,\,
1+\epsilon
\Big)\hat{A}_i
\Big).
\end{equation}

By leveraging relative advantage estimation within each group, GRPO provides a robust optimization for sequence-level reward learning.

In this work, we focus on the design of the reward function $R(q, o)$ while keeping the GRPO
optimization mechanism unchanged.
We decompose the reward into a task correctness term and an efficiency-related term based on entropy:
\begin{equation}
\label{eq:reward_decomposition}
R(q, o)
=
R_{\mathrm{corr}}(q, o)
\;\triangleright\;
\lambda \, R_{\mathrm{entropy}}(o),
\end{equation}
where $R_{\mathrm{corr}}$ evaluates the correctness of the final answer, and
$R_{\mathrm{entropy}}$ provides feedback on the reasoning trajectory. $\triangleright$ denotes that the entropy reward is combined with the correctness reward.
In the following, we design $R_{\mathrm{entropy}}$ to encourage early termination once
sufficient information has been acquired, leading to concise yet effective reasoning.




\section{Motivation}\label{sec:motivation}

Recent studies~\cite{huang2025pear,step_entropy} show that higher entropy in chain-of-thought (CoT) reasoning correlates with longer reasoning traces. This motivates reducing entropy via supervised fine-tuning or reinforcement learning to improve efficiency. However, these approaches implicitly assume that low entropy is desirable at every reasoning step. In practice, high-entropy steps often correspond to self-reflective reasoning, as illustrated in Figure~\ref{fig:two_column} (left). As a result, directly minimizing overall CoT entropy may discourage self-reflection altogether.

This assumption contradicts the nature of human reasoning.
Early stages of complex reasoning are often exploratory and uncertain, while later stages become increasingly focused and deterministic.
Efficient reasoning is therefore characterized not by uniformly low uncertainty, but by a progressive narrowing of plausible solutions.

To validate this intuition, we analyze step-wise entropy dynamics in generated CoTs on the MATH500 dataset \cite{hendrycks2021math}.
Specifically, we quantify whether uncertainty evolution along a CoT is dominated by entropy descent or ascent using the Spearman rank correlation ($\rho$) between the reasoning step index and the entropy at each step\footnote{We show the detailed experimental setup in Appendix~\ref{apx:motivation}.}. A smaller $\rho$ indicates a stronger global entropy descent.
Figure~\ref{fig:two_column} (right) reveals a clear monotonic relationship between $\rho$ and reasoning length. As $\rho$ increases, the generated token length grows correspondingly, indicating that weaker global entropy descent is associated with longer reasoning trajectories, whereas CoTs with more pronounced entropy reduction (smaller $\rho$) tend to terminate earlier. This empirical observation suggests that reasoning efficiency is closely tied to the global entropy trend.

\section{Method}
In this section, we formalize Entropy Trend Reward (ETR), a trajectory-level reward for efficient chain-of-thought (CoT) reasoning.
ETR is designed to encourage progressive uncertainty reduction along a CoT while preserving correctness.

\subsection{Entropy Trajectory of Chain-of-Thought}

Given an input question $q$, the model generates a chain-of-thought response
\[
o = \{C_1, C_2, \ldots, C_T\},
\]
where each $C_t$ corresponds to one reasoning step. In practice, we split reasoning steps using the “\textbackslash n\textbackslash n” delimiter.

For each step $t$, we define a scalar entropy measure
\begin{equation}
H_t = H\!\left(p_\theta(\cdot \mid C_{1:t})\right),
\end{equation}
where $H(\cdot)$ denotes the Shannon entropy of the model’s next-token predictive distribution.
This provides a model-internal estimate of predictive uncertainty at each reasoning step.

We define the step-wise entropy change as:
\begin{equation}
\Delta_t = H_{t-1} - H_t, \qquad t = 2,\ldots,T.
\label{eq:entropy_delta}
\end{equation}
A positive $\Delta_t$ indicates uncertainty reduction, while a negative value corresponds to increased uncertainty or exploratory divergence.




\begin{figure*}[!t]
\centering
\centering
\includegraphics[width=1\linewidth]{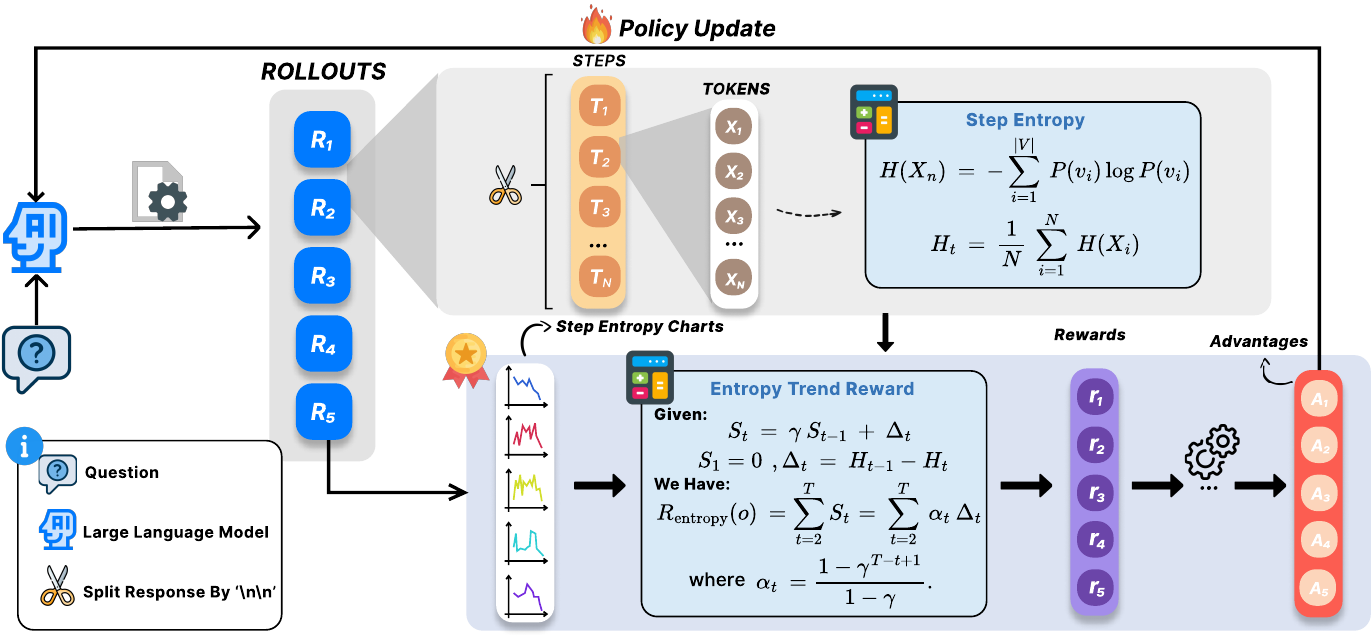}
\caption{Overview of Entropy Trend Reward (ETR) in RL training.
ETR computes step-wise entropy from generated rollouts, aggregates entropy changes with momentum to capture global entropy descent, and provides a reward signal, combined with correctness supervision, to guide policy updates toward more convergent reasoning.}
\label{fig:method_overview}
\end{figure*}

\begin{figure}[!t]
\centering
\includegraphics[width=1\linewidth]{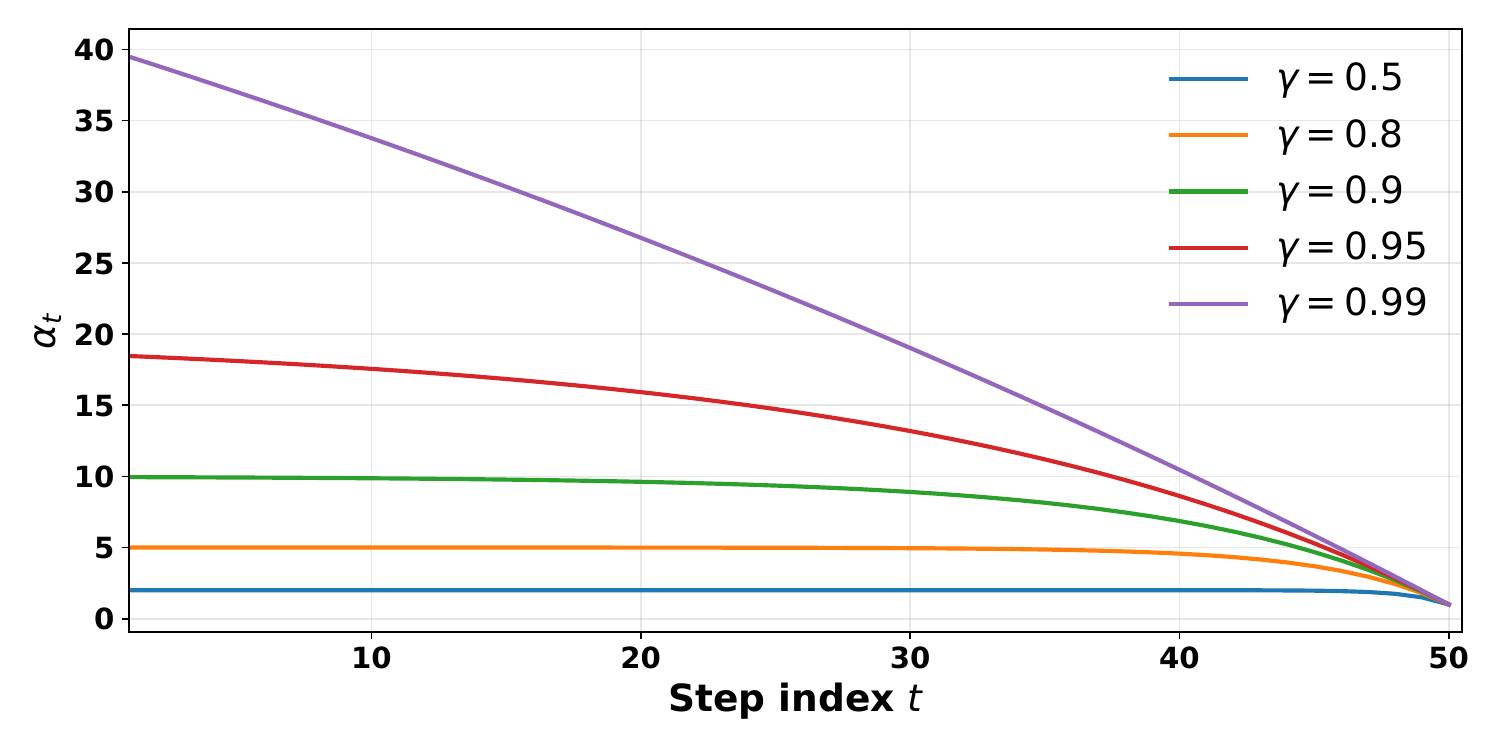}
\caption{Evolution of the cumulative momentum weights $\alpha_t$ under different momentum coefficients $\gamma$.}
\vspace{-1em}
\label{fig:gamma}
\end{figure}

\subsection{Momentum-Based Entropy Trend Reward} \label{sec:momentum-ETR}
A natural baseline objective is to reward total entropy reduction. However, this reward telescopes:
\begin{equation}
R_{\text{naive}}(o) = \sum_{t=2}^{T} \Delta_t = H_1 - H_T,
\end{equation}
and thus depends only on the initial and final entropy values.
As a consequence, fundamentally different entropy trajectories, e.g., smooth monotonic descent versus repeated entropy spikes, receive identical rewards as long as they share the same endpoints.


To address this, we introduce a momentum-based accumulation of entropy changes.
We define a latent trend variable $S_t$ recursively:
\begin{equation}
S_t = \gamma S_{t-1} + \Delta_t, \qquad S_1 = 0,
\label{eq:momentum}
\end{equation}
where $\gamma \in (0,1)$ is a momentum coefficient controlling temporal smoothing.

The entropy reward for a CoT is then defined as
\begin{equation}
R_{\mathrm{entropy}}(o)
= \sum_{t=2}^{T} S_t.
\label{eq:entropy_reward}
\end{equation}

Unrolling the recurrence yields
\begin{align}
R_{\mathrm{entropy}}(o)
&= \sum_{t=2}^{T} \alpha_t \, \Delta_t, \\
\text{where } \alpha_t
&= \frac{1 - \gamma^{T - t + 1}}{1 - \gamma}.
\label{eq:alpha}
\end{align}

Unlike the naive entropy trend reward, which ignores intermediate reasoning structure, the momentum-based formulation assigns a gradient signal to \emph{every} step in the trajectory:
Thus, each entropy drop (or increase) influences the reward, providing fine-grained feedback that guides the policy toward coherent and efficient reasoning.

A key property of $\alpha_t$ is that it is strictly decreasing in $t$.   To illustrate, we show an example plot of $\alpha_t$ in Figure \ref{fig:gamma}. Because $\alpha_t$ decreases as $t$ increases, reductions in uncertainty that occur earlier in the reasoning trajectory are rewarded more than those occurring later. This encourages the model to converge its belief state rapidly rather than postponing critical deductions to late steps. In our experiments, we set $\gamma = 0.9$, with a detailed analysis of its selection deferred to Appendix~\ref{apx:momentum}.

\subsection{Instance-Adaptive Termination via Entropy Trend Shaping}

Our reward shaping implicitly encourages instance-adaptive termination without imposing any explicit length constraint.
Recall that ETR maintains a momentum-aggregated trend state
$S_t=\gamma S_{t-1}+\Delta_t$ with $\Delta_t=H_{t-1}-H_t$.
Since the entropy reward accumulates $\sum_{t=2}^{T} S_t$, generating additional steps is beneficial only when the next update keeps the trend state positive, i.e., when the next step continues to reduce uncertainty on average.
In contrast, steps that increase entropy ($\Delta_t<0$) decrease $S_t$ and are repeatedly penalized, which suppresses oscillatory self-reflection loops.

This mechanism yields instance-adaptive behavior.
For \textbf{easy instances}, entropy typically drops early, leading to large positive $\Delta_t$ and a rapid rise of $S_t$; after the model becomes confident, further steps rarely provide additional entropy descent and are therefore disfavored, resulting in short CoTs.
For \textbf{hard instances}, entropy may fluctuate; ETR discourages large entropy ascents and favors trajectories with steady, incremental uncertainty reduction, allocating more reasoning steps only when they contribute to convergence.
Compared with global entropy penalties~\cite{huang2025pear,step_entropy}, ETR is trajectory-aware: it tolerates limited local exploration while enforcing a coherent global descent.

\subsection{Integration with GRPO}

We incorporate the entropy trend reward into GRPO by decomposing the overall reward as
\begin{equation}
R(q, o)
=
R_{\mathrm{corr}}(q, o)
\;\triangleright\;
\lambda R_{\mathrm{entropy}}(o),
\label{eq:final_reward}
\end{equation}
where $R_{\mathrm{corr}}$ evaluates the correctness of the final answer. The operator $\triangleright$ now represents that the entropy reward is
applied only as a shaping term for correct trajectories. And $\lambda > 0$ controls the influence of entropy-based efficiency shaping.

In practice, we adopt a two-stage reward structure:
\begin{equation}
R(q, o)
=
\begin{cases}
-1, & \text{if incorrect}, \\[4pt]
1 + \lambda R_{\mathrm{entropy}}(o), & \text{if correct}.
\end{cases}
\end{equation}
An overview of Entropy Trend Reward (ETR) in RL training is illustrated in Figure~\ref{fig:method_overview}. And the detailed procedure is summarized in Appendix~\ref{apx:algorithm},  This design ensures that correctness is a hard constraint, while entropy shaping only affects the correct reasoning trajectories.
Within each rollout group, GRPO’s relative advantage normalization further emphasizes comparative efficiency among valid CoTs generated for the same input.

%% file: experiments.tex
\section{Experiments and Analysis}
\subsection{Experimental Setup}

\noindent
\textbf{Dataset.}  
Our training data consists of 7,000 problems randomly sampled from DeepMath-103K \citep{deepmath}, covering difficulty levels from 5 to 10 only. We additionally perform a verification step to ensure that the sampled training data has no overlap with the held-out evaluation benchmarks.

\noindent
\textbf{Benchmarks.} To comprehensively evaluate the reasoning ability of our model, we adopt both mathematical and general reasoning benchmarks. For math-specific evaluation, we select AIME24 \citep{aime24}, AMC23 \citep{amc23}, and MATH500 \citep{hendrycks2021math}, which emphasize high-school to competition-level mathematical problem solving. For broader knowledge-intensive reasoning, we adopt GPQA Diamond \citep{rein2023gpqa}, a subset of the GPQA benchmark that focuses on expert-level question answering. 

\noindent
\textbf{Reasoning Models.} We evaluate our approach using three reasoning models: DeepSeek-R1-Distill-Qwen-7B\footnote{\url{https://huggingface.co/deepseek-ai/DeepSeek-R1-Distill-Qwen-7B}}, Qwen3-4B\footnote{\url{https://huggingface.co/Qwen/Qwen3-4B}}, Qwen3-8B\footnote{\url{https://huggingface.co/Qwen/Qwen3-8B}}. These models generally exhibit strong reasoning abilities but with excessive output tokens.

\definecolor{blue}{RGB}{225,239,253}   

\begin{table*}[t]
\centering
\scriptsize
\setlength{\tabcolsep}{5.2pt}
\begin{tabular}{l|cccccccccccc|ccc}
\toprule
\multirow{3}{*}{\textbf{Method}} &
\multicolumn{3}{c}{\textbf{AMC23}} &
\multicolumn{3}{c}{\textbf{AIME24}} &
\multicolumn{3}{c}{\textbf{MATH500}} &
\multicolumn{3}{c}{\textbf{GPQA-D}} &
\multicolumn{3}{c}{\textbf{Overall}} \\
\cmidrule(lr){2-4}\cmidrule(lr){5-7}\cmidrule(lr){8-10}\cmidrule(lr){11-13}\cmidrule(lr){14-16}
&
\textbf{Acc}$\uparrow$ & \textbf{Len}$\downarrow$ & \textbf{CR}$\downarrow$ &
\textbf{Acc}$\uparrow$ & \textbf{Len}$\downarrow$ & \textbf{CR}$\downarrow$ &
\textbf{Acc}$\uparrow$ & \textbf{Len}$\downarrow$ & \textbf{CR}$\downarrow$ &
\textbf{Acc}$\uparrow$ & \textbf{Len}$\downarrow$ & \textbf{CR}$\downarrow$ &
\textbf{Acc}$\uparrow$ & \textbf{Len}$\downarrow$ & \textbf{AES}$\uparrow$ \\

\midrule
\rowcolor{gray!15}
\multicolumn{16}{l}{\textit{\textbf{DeepSeek-R1-Distill-7B}}} \\
Original & 
80.0 & 6.6k & 100\% &
43.3 & 11.8k & 100\% &
85.0 & 4.2k & 100\% &
24.2 & 11.3k & 100\% &
58.1 & 8.5k & 0.00 \\
DEER &
85.0 & 4.9k & 74.2\% &
50.0 & 9.9k & 83.9\% &
85.8 & 2.3k & 54.8\% &
22.7 & 7.6k & 67.3\% &
60.9 & 6.2k & 0.51 \\
NoThink &
77.5 & 3.8k & 57.6\% &
56.7 & 8.3k & 70.3\% &
81.4 & 1.7k & 40.5\% &
22.2 & 2.0k & 17.7\% &
59.5 & 4.0k & 0.65 \\
LCPO & 
87.5 & 3.5k & 53.0\% & 50.0 & 6.8k & 57.8\% & 85.4 & 2.2k & 52.6\% & 11.61 & 2.8k & 24.6\% & 58.6 & 3.8k & 0.60 \\
O1-Pruner & 
92.5 & 3.2k & 48.4\% & 
46.7 & 7.8k & 66.1\% & 
89.0 & 2.1k & 50.0\% & 
39.4 & 6.2k & 54.9\% &
66.9 & 4.8k & 1.18 \\

PEAR & 92.5 & 4.4k & 66.04\% & 60.0 & 8.5k & 72.2\% & 90.2 & 2.5k & 59.9\% & 36.9 & 5.2k & 45.9\% & 69.8 & 5.1k & 1.41\\
\rowcolor{blue}
\textbf{ETR} &
\textbf{87.5} & \textbf{2.4k} & \textbf{36.4\%} &
\textbf{56.7} & \textbf{4.6k} & \textbf{39.0\%} &
\textbf{90.6} & \textbf{1.5k} & \textbf{35.7\%} &
\textbf{37.3} & \textbf{2.5k} & \textbf{22.1\%} &
\textbf{68.0} & \textbf{2.8k} & \textbf{1.53} \\
\midrule
\rowcolor{gray!15}
\multicolumn{16}{l}{\textit{\textbf{Qwen3-4B}}} \\
Original & 
90.0 & 7.6k & 100\% &
53.3 & 11.7k & 100\% &
90.6 & 5.0k & 100\% &
43.9 & 10.4k & 100\% &
69.5 & 8.7k & 0.00 \\
DEER &
92.5 & 6.1k & 80.3\% &
56.7 & 11.5k & 98.3\% &
92.2 & 3.7k & 74.0\% &
52.5 & 8.2k & 78.8\% &
73.5 & 7.4k & 0.44 \\
NoThink &
82.5 & 2.4k & 31.6\% &
33.3 & 8.0k & 68.4\% &
85.6 & 1.7k & 34.0\% &
23.2 & 4.7k & 45.2\% &
56.2 & 4.2k & -0.44 \\
LCPO &
90.0 & 6.0k & 78.9\% &
50.0 & 8.2k & 70.6\% &
90.8 & 4.5k & 89.2\% &
47.5 & 3.8k & 36.3\% &
69.6 & 5.6k & 0.36 \\
O1-Pruner & 
85.0 & 2.6k & 34.2\% & 
50.0 & 6.9k & 59.0\% & 
90.6 & 1.8k & 36\% & 
50.0 & 3.3k & 31.7\% &
68.0 & 3.6k & 0.54 \\
PEAR & 
92.5 & 6.0k & 34.2\% & 
70.0 & 9.9k & 59.0\% & 
91.8 & 3.8k & 36\% & 
54.54 & 7.1k & 31.7\% &
77.2 & 6.7k & 0.79 \\
\rowcolor{blue}
\textbf{ETR} &
\textbf{90.0} & \textbf{4.0k} & \textbf{52.6\%} &
\textbf{73.3} & \textbf{7.5k} & \textbf{64.1\%} &
\textbf{91.4} & \textbf{2.1k} & \textbf{42.0\%} &
\textbf{53.5} & \textbf{4.1k} & \textbf{39.4\%} &
\textbf{77.1} & \textbf{4.4k} & \textbf{1.03} \\
\midrule
\rowcolor{gray!15}
\multicolumn{16}{l}{\textit{\textbf{Qwen3-8B}}} \\
Original & 
90.0 & 8.0k & 100\% &
63.3 & 12.2k & 100\% &
90.6 & 5.4k & 100\% &
52.0 & 9.9k & 100\% &
74.0 & 8.9k & 0.00 \\
DEER &
92.5 & 6.4k & 80.0\% &
63.3 & 10.3k & 84.4\% &
93.0 & 3.1k & 57.4\% &
61.6 & 9.0k & 90.9\% &
77.6 & 7.2k & 0.43  \\
NoThink &
67.5 & 3.4k & 41.3\% &
40.0 & 7.0k & 57.4\% &
86.4 & 1.5k & 27.8\% &
26.8 & 4.4k & 44.4\% &
55.2 & 4.1k & -0.73  \\
LCPO &
90.0 & 5.7k & 41.3\% &
60.0 & 7.0k & 57.4\% &
93.6 & 4.8k & 27.8\% &
54.6 & 4.0k & 44.4\% &
74.5 & 5.4k & 0.43  \\
O1-Pruner & 
90.0 & 3.3k & 41.3\% & 
66.7 & 6.6k & 54.1\% & 
92.2 & 1.9k & 35.2\% & 
56.1 & 4.1k & 41.4\% &
76.3 & 4.0k & 0.70  \\

PEAR & 
87.5 & 6.8k & 84.7\% & 
63.3 & 11.0k & 90.1\% & 
92.4 & 4.5k & 83.5\% & 
55.1 & 8.2k & 83.3\% &
74.6 & 7.6k & 0.18  \\
\rowcolor{blue}
\textbf{ETR} &
\textbf{92.5} & \textbf{4.2k} & \textbf{53.8\%} &
\textbf{73.3} & \textbf{8.6k} & \textbf{75.4\%} &
\textbf{93.6} & \textbf{2.4k} & \textbf{44.4\%} &
\textbf{57.1} & \textbf{5.2k} & \textbf{52.5\%} &
\textbf{79.1} & \textbf{5.1k} & \textbf{0.77} \\
\bottomrule
\end{tabular}
\vspace{-0.5em}
\caption{Evaluation results on four benchmarks using greedy decoding (pass@1). \textbf{Acc} denotes accuracy, \textbf{Len} denotes token length, and \textbf{CR} denotes compression rate. $\uparrow$ and $\downarrow$ indicate higher- and lower-is-better, respectively. \textbf{AES} is a unified score that jointly accounts for accuracy and efficiency.
} \label{tab:main_table}
\vspace{-0.5em}
\end{table*}

\noindent
\textbf{Baselines.} 
We compare our method with several representative approaches for efficient reasoning, including both training-free and RL–based methods. Training-free methods such as DEER \citep{yang2025dynamic} and NoThink \citep{nothinking} reduce response length through prompt  or early stopping criteria design.  RL–based approaches, including LCPO \citep{L1}, O1-Pruner \citep{O1-pruner}, and PEAR \citep{huang2025pear}, address this problem by designing reward functions that explicitly depend on sequence length or entropy.

\noindent
\textbf{Evaluation and Metrics.} 
In this work, we evaluate our method using accuracy (Acc), response length (Len), compression rate (CR), and the Accuracy–Efficiency Trade-off Score (AES) \citep{O1-pruner}. CR is defined as the ratio of the average response length to that of the original model, where lower values indicate higher compression. AES measures the trade-off between efficiency gains and accuracy changes relative to a baseline.

\begin{equation*}
\begin{aligned}
\mathrm{AES} &=
\underbrace{\frac{L_{\text{base}} - L_{\text{model}}}{L_{\text{base}}}}_{\text{relative length reduction}}
\;+\;
\varsigma \,
\underbrace{
\frac{A_{\text{model}} - A_{\text{base}}}{A_{\text{base}}}
}_{\text{relative accuracy change}},
\end{aligned}
\end{equation*}
where 
$L$ and 
$A$ denote response length and accuracy, and 
$\varsigma$ balances accuracy improvement and efficiency gains. Following \citet{O1-pruner}, we set $\varsigma=5$ to prioritize accuracy over length.


\noindent
\textbf{Training Details} We conduct training using the open-source VeRL framework \citep{sheng2024hybridflow} on 8 NVIDIA H-100 GPUs. To reduce memory footprint and improve training efficiency, we employ Low-Rank Adaptation (LoRA) \citep{hu2021lora}. Unless otherwise specified, we use a batch size of 32 and a learning rate of $1 \times 10^{-5}$. The maximum response length is set to $16,384$, with 5 rollouts per sample.

\subsection{Main Results}

Table \ref{tab:main_table} reports the accuracy–efficiency trade-off of different methods across multiple reasoning benchmarks and model configurations. We highlight two consistent observations.

First, ETR performs robustly across different model families and sizes. Across both DeepSeek-R1-Distill and Qwen3 model families, spanning sizes from 4B to 8B, our method consistently maintains strong accuracy while substantially reducing reasoning length, achieving the highest AES scores in all settings. This indicates that ETR does not rely on a specific model architecture or scale, but generalizes well across comparable model sizes and across different model families.

Second, and more importantly, ETR achieves the best balance between accuracy and reasoning length among all compared methods.
Existing training-free approaches (e.g., DEER and NoThink) struggle to jointly optimize these two objectives. DEER largely preserves accuracy but fails to substantially reduce output length, while NoThink aggressively truncates reasoning traces (e.g., on Qwen3-4B, reducing output from 8.7k to 4.2k tokens), but at the cost of severe accuracy degradation (69.5\% → 56.2\%), resulting in negative AES scores.
Methods based on explicit length penalties (e.g., LCPO and O1-Pruner) shorten outputs more effectively than training-free approaches, but this is often accompanied by significant accuracy loss. Entropy-based reinforcement learning methods (e.g., PEAR) tend to preserve accuracy, yet still produce long reasoning chains due to their reliance on global entropy signals.

In contrast, ETR consistently yields the highest AES across all evaluated settings, demonstrating a more principled control of the reasoning process that avoids both underthinking and overthinking. 

\subsection{Empirical Analysis}
Unless otherwise specified, we use DeepSeek-R1-Distill-Qwen-7B as the base model for all empirical analyses in this section.

\noindent
\textbf{6.3.1 Analysis on Entropy Rewards.}

We conduct an ablation study on entropy-based reward designs to isolate the effects of entropy magnitude, temporal aggregation, and correctness supervision. Results are summarized in Table~\ref{tab:entropy_ablation}.

\textbf{Absolute entropy objectives.}
We first examine two extreme strategies that directly optimize the average predictive entropy over all CoT tokens.
Entropy minimization (Min. $H$), as adopted in prior work~\cite{step_entropy,huang2025pear}, yields short CoTs but does not improve accuracy and achieves a lower AES score than our method (1.06 vs.\ 1.53).
In contrast, entropy maximization (Max. $H$) collapses reasoning, producing maximal-length generations with near-zero accuracy.
These results show that optimizing absolute entropy magnitude alone is suboptimal for efficient reasoning.

\textbf{Effect of momentum.}
Removing temporal aggregation (No $\gamma$) corresponds to the naive variant discussed at the beginning of Section~\ref{sec:momentum-ETR}, which relies only on entropy differences between first and last sentences.
As a result, it fails to capture the overall entropy trend along the CoT.
While accuracy remains reasonable because of correctness reward, it cannot reduce CoT length, highlighting the necessity of momentum-based aggregation.


\textbf{Effect of correctness supervision.}
Removing the correctness reward (No $R_{\mathrm{corr}}$) leads to short CoTs but causes a substantial accuracy drop (e.g., 65.0\% on AMC23), indicating that entropy trends alone are insufficient for ensuring correctness.
Overall, combining entropy trends with momentum-based entropy trend reward and correctness supervision yields the best accuracy--efficiency trade-off across all benchmarks.

\begin{table}[t]
\centering
\scriptsize
\setlength{\tabcolsep}{1.8pt}
\begin{tabular}{lccccccccl}
\toprule
\multirow{2}{*}{\textbf{Reward}} & \multicolumn{2}{c}{\textbf{AMC23}} & \multicolumn{2}{c}{\textbf{AIME24}} & \multicolumn{2}{c}{\textbf{MATH500}} & \multicolumn{2}{c}{\textbf{GPQA-D}} &\multirow{2}{*}{\textbf{AES}$\uparrow$}\\
\cmidrule(lr){2-3} \cmidrule(lr){4-5} \cmidrule(lr){6-7} \cmidrule(lr){8-9}
& Acc $\uparrow$ & Len $\downarrow$ & Acc $\uparrow$ & Len $\downarrow$ & Acc $\uparrow$ & Len $\downarrow$ & Acc $\uparrow$ & Len $\downarrow$ \\
\midrule
Original & 
80.0 & 6.6k  &
43.3 & 11.8k  &
85.0 & 4.2k  &
24.2 & 11.3k & 0.00  \\ \midrule
Min. $H$ & 80.0 & 2.1k & 43.3 & 5.1k & 88.2 & 1.3k & 38.3 & 2.1k & 1.06\\
Max. $H$ & 10.0 & 15.1k & 0.0 & 16.4k & 9.0 & 15.3k & 1.5 & 16.0k & -5.4 \\
\midrule
No $\gamma$ & 87.5 & 4.9k & 46.67 & 10.0k & 87.8 & 3.6k & 31.8 & 10.0k & 0.61 \\
No $R_{\mathrm{corr}}$ & 65.0 & 1.2k & 23.3 & 1.4k & 78.6 & 0.7k & 29.8 & 0.7k & 0.11\\
\midrule
\textbf{Ours} & \textbf{87.5} & \textbf{2.4k} & \textbf{56.7} & \textbf{4.6k} & \textbf{90.6} & \textbf{1.5k} & \textbf{37.4} & \textbf{2.5k} & \textbf{1.53} \\
\bottomrule
\end{tabular}
\caption{Ablation Study on Entropy-based Reward Designs. We compare our momentum-based reward against extreme constraints and simplified variants across four benchmarks. Our method demonstrates a superior balance between accuracy and efficiency.}
\label{tab:entropy_ablation}
\end{table}

\noindent
\textbf{6.3.3 Analysis on Entropy Trend.}
\begin{figure}[!t]
    \centering
    \includegraphics[width=1\linewidth]{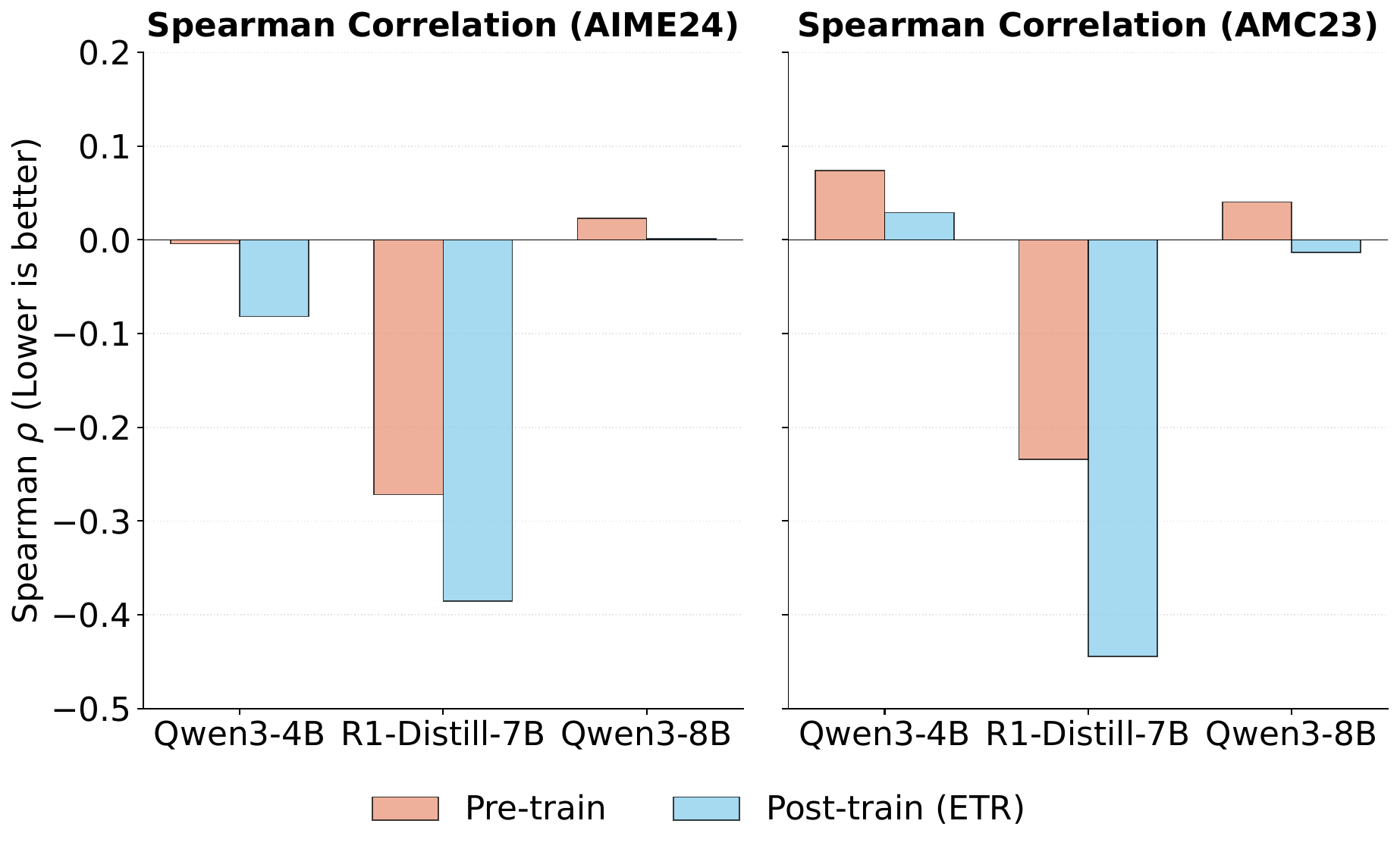}
    \vspace{-2em}
    \caption{Spearman's rank correlation coefficient between reasoning steps and step entropy. The shift toward negative values indicates a more convergent reasoning trajectory after ETR training.}
    \label{fig:entropy_correlation}
\end{figure}

To assess whether Entropy Trend Reward (ETR) promotes convergent reasoning, we analyze entropy dynamics along the Chain-of-Thought (CoT). 
We compute the Spearman rank correlation ($\rho$) between the reasoning step index and the entropy at each step, where a more negative $\rho$ indicates a stronger global entropy descent.

As shown in Figure~\ref{fig:entropy_correlation}, across model scales, ETR consistently shifts the correlation from positive or near-zero to negative values. This sign reversal provides empirical evidence that ETR effectively regularizes the reasoning path, ensuring that each step contributes to narrowing the solution space rather than accumulating logical ambiguity.

\noindent

\textbf{6.3.5 Behavioral View of CoT Reduction.}
\begin{figure}[!t]
    \centering
    \includegraphics[width=1\linewidth]{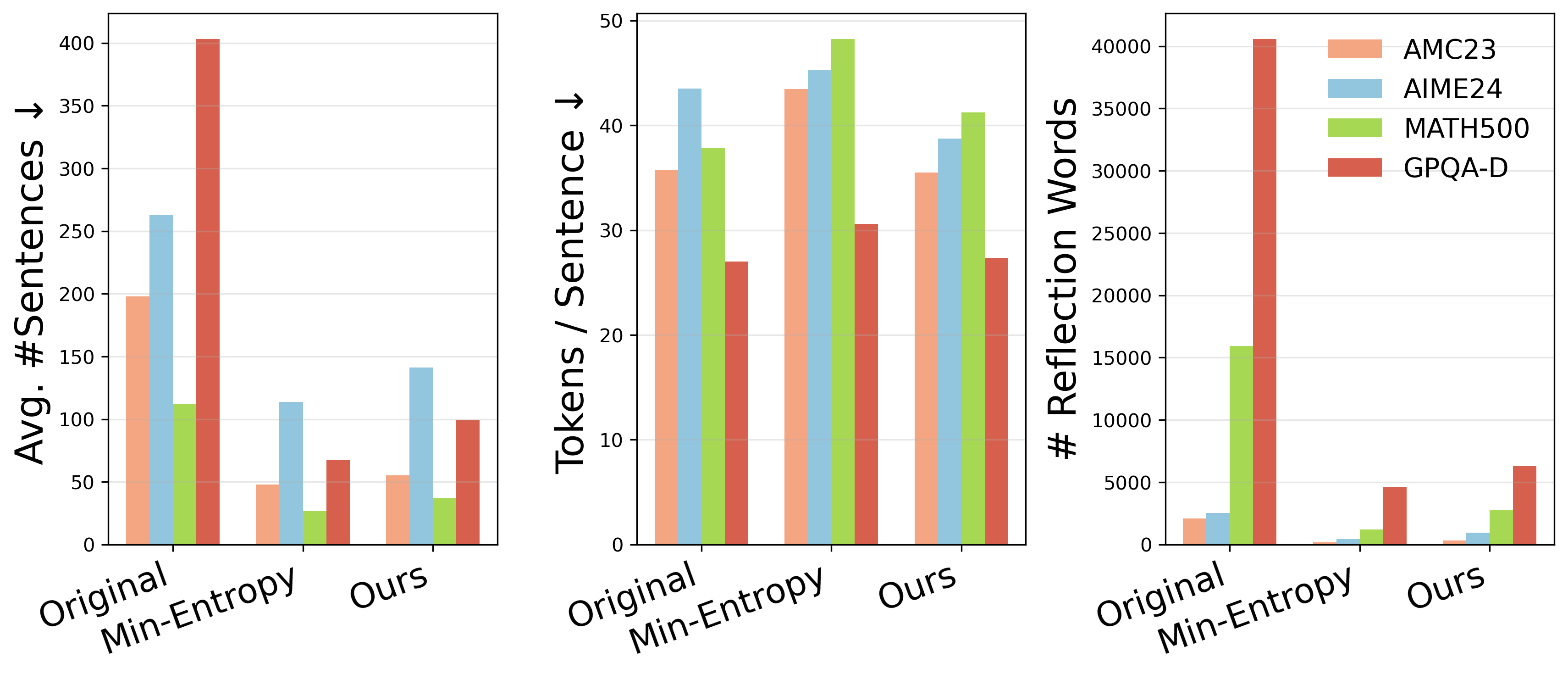}
    \vspace{-2em}
    \caption{ETR reduces CoT length primarily by maintaining low per-step verbosity, rather than by aggressively suppressing reasoning steps or reflective processes, in contrast to entropy minimization.}
    \label{fig:behave_view}
\end{figure}


Figure~\ref{fig:behave_view} decomposes CoT length into (i) the number of sentence-level steps, (ii) tokens per step, and (iii) the number of self-reflection markers\footnote{Self-reflection words: \emph{wait, alternatively, hmm, but, however, alternative, another, check, double-check, oh, maybe}.}.
Direct entropy minimization shortens CoTs primarily by suppressing exploration, yielding fewer steps and dramatically fewer reflection markers, but at the cost of more verbose reasoning steps (higher tokens per sentence).
In contrast, ETR preserves a moderate degree of self-verification and multi-step reasoning while maintaining low per-step verbosity, aligning with our objective of permitting limited local exploration under a globally convergent uncertainty trajectory.
This difference explains why ETR achieves a superior accuracy–efficiency trade-off: it discourages oscillatory overthinking without enforcing uniformly low uncertainty at every step.



%% file: conclusion.tex
\section{Conclusion}
In this paper, we show that chain-of-thought efficiency is governed by the global trend of predictive uncertainty rather than its absolute magnitude. Based on this insight, we propose Entropy Trend Reward (ETR), which encourages progressive uncertainty reduction while enforcing correctness as a hard constraint. Extensive experiments demonstrate that ETR consistently improves the accuracy--efficiency trade-off.

\section*{Limitation.}
Owing to computational constraints, our comparison with state-of-the-art methods is limited to models up to 8B parameters using LoRA-based training. We plan to scale our method to larger models and perform comparisons in future work.

%% file: appendix.tex
\section{Algorithmic Details of Entropy Trend Reward}\label{apx:algorithm}
Algorithm~\ref{alg:entropy_reward} presents the detailed computation of Entropy Trend Reward (ETR) for a generated chain-of-thought.
Given a sentence-level CoT response, the algorithm first enforces correctness as a hard constraint by assigning a negative reward to incorrect outputs.
For correct responses, it computes step-wise predictive entropy, aggregates entropy changes using a momentum-based recurrence, and accumulates a trajectory-level entropy reward.
The final reward combines correctness supervision with entropy-based efficiency shaping, providing dense feedback that encourages progressive uncertainty reduction during reasoning.

\begin{algorithm}[h]
\caption{Entropy Trend Reward (ETR)}
\label{alg:entropy_reward}
\DontPrintSemicolon

\KwIn{CoT response $o=\{C_t\}_{t=1}^{T}$;\ ground-truth $y^*$;\ momentum $\gamma$;\ weight $\lambda$.}
\KwOut{Reward $R(q,o)$.}

$\hat{y} \gets \textsc{ExtractAnswer}(o)$\;
\If{$\hat{y} \neq y^*$}{\Return $-1$\;}

$H_1 \gets H\!\left(p_\theta(\cdot \mid C_{1:1})\right)$\;
$S \gets 0;\ R_{\mathrm{entropy}} \gets 0$\;

\For{$t \gets 2$ \KwTo $T$}{
  $H_t \gets H\!\left(p_\theta(\cdot \mid C_{1:t})\right)$\;
  $\Delta_t \gets H_{t-1} - H_t$\;
  $S \gets \gamma S + \Delta_t$\;
  $R_{\mathrm{entropy}} \gets R_{\mathrm{entropy}} + S$\;
}

\Return $1 + \lambda R_{\mathrm{entropy}}$\;
\end{algorithm}

\section{Derivation of Equation~\ref{eq:entropy_reward}.}
Recall that the step-wise entropy change is
\begin{equation}
\Delta_t = H_{t-1} - H_t, \qquad t = 2,\ldots,T,
\label{eq:entropy_delta_app}
\end{equation}
and the momentum recursion is
\begin{equation}
S_t = \gamma S_{t-1} + \Delta_t, \qquad S_1 = 0,
\label{eq:momentum_app}
\end{equation}
with $\gamma \in (0,1)$. The entropy reward is defined as
\begin{equation}
R_{\mathrm{entropy}}(o) = \sum_{t=2}^{T} S_t .
\label{eq:entropy_reward_app}
\end{equation}

\medskip
\noindent\textbf{Step 1: Unroll the recursion.}
For any $t \ge 2$, repeatedly substituting Eq.~\eqref{eq:momentum_app} gives
\begin{align}
S_t
&= \gamma S_{t-1} + \Delta_t \nonumber\\
&= \gamma(\gamma S_{t-2} + \Delta_{t-1}) + \Delta_t \nonumber\\
&= \gamma^2 S_{t-2} + \gamma \Delta_{t-1} + \Delta_t \nonumber\\
&\ \ \vdots \nonumber\\
&= \gamma^{t-2} S_2 + \sum_{k=3}^{t} \gamma^{t-k}\Delta_k . \label{eq:unroll_mid}
\end{align}
Since $S_2=\gamma S_1+\Delta_2=\Delta_2$ and $S_1=0$, Eq.~\eqref{eq:unroll_mid} simplifies to
\begin{equation}
S_t = \sum_{k=2}^{t} \gamma^{\,t-k}\,\Delta_k .
\label{eq:unroll_closed}
\end{equation}

\medskip
\noindent\textbf{Step 2: Substitute into the reward and swap summations.}
Plugging Eq.~\eqref{eq:unroll_closed} into Eq.~\eqref{eq:entropy_reward_app} yields
\begin{align}
R_{\mathrm{entropy}}(o)
&= \sum_{t=2}^{T} \sum_{k=2}^{t} \gamma^{\,t-k}\,\Delta_k \nonumber\\
&= \sum_{k=2}^{T} \Delta_k \sum_{t=k}^{T} \gamma^{\,t-k} ,
\label{eq:swap_sum}
\end{align}
where the last line follows by exchanging the order of summation.

\medskip
\noindent\textbf{Step 3: Evaluate the inner geometric series.}
Let $m=t-k$, so that $m=0,\ldots,T-k$. Then
\begin{equation}
\sum_{t=k}^{T} \gamma^{\,t-k}
= \sum_{m=0}^{T-k} \gamma^{\,m}.
\end{equation}
For $\gamma\neq 1$, this is a finite geometric series:
\begin{equation}
\sum_{m=0}^{T-k} \gamma^{\,m}
= \frac{1-\gamma^{\,T-k+1}}{1-\gamma}.
\label{eq:geom_sum}
\end{equation}
Substituting Eq.~\eqref{eq:geom_sum} into Eq.~\eqref{eq:swap_sum} gives
\begin{equation}
R_{\mathrm{entropy}}(o)
= \sum_{k=2}^{T} \alpha_k \,\Delta_k,
\qquad
\alpha_k=\frac{1-\gamma^{\,T-k+1}}{1-\gamma}.
\end{equation}
Renaming the index $k\mapsto t$ yields Eq.~\eqref{eq:alpha} in the main text:
\begin{equation}
R_{\mathrm{entropy}}(o)
= \sum_{t=2}^{T} \alpha_t \,\Delta_t,
\qquad
\alpha_t=\frac{1-\gamma^{\,T-t+1}}{1-\gamma}.
\end{equation}

\begin{table*}[!t]
\centering

\begin{tabular}{lcccc}
\toprule
Hyperparameter & LCPO & O1-Pruner & PEAR & ETR (Ours) \\ \midrule
LoRA Rank & - & - & - & 32 \\
LoRA Alpha & - & - & - & 64 \\
Actor learning rate & $1e-6$ & $1e-6$ & $1e-6$ & $1e-5$ \\
train\_batch\_size & 64 & 32 & 32 & 32 \\
mini\_batch\_size & 64 & 16 & 16 & 16 \\
micro\_batch\_size & - & 2 & 2 & 2 \\
Training step & 1400 & 207 & 207 & 207 \\
Max response length & 4096 & 16384 & 16384 & 16384 \\
Num of rollouts & 8 & 5 & 5 & 5 \\
Rollout temp ($\tau$) & 0.6 & 1.0 & 1.0 & 1.0 \\
KL penalty ($\beta$) & $1e-3$ & $1e-3$ & $1e-3$ & $5e-3$ \\
Entropy Coef & - & - & - & $1e-4$\\
Advantage clip ($\epsilon$) & \multicolumn{4}{c}{0.2} \\

\bottomrule
\end{tabular}
\caption{Training configs for various methods.}
\label{tab:training_configs}
\end{table*}


\section{Training and Evaluation Details for ETR}
\begin{table}[H]
\centering

\begin{tabular}{lc}
\toprule
Hyperparameter & ETR (Ours) \\ \midrule
LoRA Rank & 32 \\
LoRA Alpha & 64 \\
LoRA Target Modules & all-linear \\
Actor learning rate & $1e-5$ \\
train\_batch\_size & 32 \\
mini\_batch\_size & 16 \\
micro\_batch\_size & 2 \\
param\_offload & True \\
optimizer\_offload & True \\
log\_prob\_micro\_batch\_size & 8 \\
enable\_chunked\_prefill & True \\
Training step & 207 \\
Max response length & 16384 \\
Num of rollouts & 5 \\
Rollout temp ($\tau$) & 1.0 \\
KL penalty ($\beta$) & $5e-3$ \\
Entropy Coef & $1e-4$\\
Advantage clip ($\epsilon$) & 0.2 \\
epochs & 1 \\
\bottomrule
\end{tabular}
\caption{Detaied Configurations for ETR}
\label{tab:training_config_ours}
\end{table}

During training and evaluation, we insert the below system prompt before the question:
\newtcolorbox{sys_prompt}{colback=white!5, colframe=white!60!black, title=System Prompt , breakable}
\begin{sys_prompt}
"Please reason step by step, and put your final answer within \verb|\boxed{}|"
\end{sys_prompt}

Our evaluation methodology utilizes the SGlang framework \citep{zheng2024sglang} to conduct high-throughput inference across multiple benchmarks, including MATH500 \citep{hendrycks2021math}, GPQA Diamond \citep{rein2023gpqa}, AIME24 \citep{aime24}, and AMC23 \citep{amc23}. We configured the generation process with greedy decoding, a max response length of 16,384 and a sample size of 1, ensuring the model has sufficient space to develop complex reasoning paths while maintaining deterministic outputs for accuracy tracking. To extract the final answer from the model's response, we leveraged implemented functions in VeRL library \citep{sheng2024hybridflow}. This process specifically extracts the final answer in \verb|\boxed{}|.

For the final assessment of correctness, we integrated the grading logic from the OpenAI PRM800K dataset grader \citep{lightman2023let}. We found this approach to be significantly reliable and robust, as it effectively handles differing mathematical formulations and varied formatting styles. Throughout the evaluation, we recorded both the accuracy and the average response length for each dataset, allowing us to analyze the relationship between reasoning efficiency and model performance.

\section{Training Details For Baseline Methods}
 We selected five baseline methods for comparison: \textbf{DEER} \citep{yang2025dynamic}, \textbf{NoThink} \citep{nothinking}, \textbf{LCPO} \citep{L1}, \textbf{O1-Pruner} \citep{O1-pruner} and \textbf{PEAR} \citep{huang2025pear}. Among these, DEER and NoThink are training-free approaches, while the remaining methods are trained using the DeepMath-103K dataset \citep{deepmath}. The training details can be found below. We report all training hyperparameters in Table \ref{tab:training_configs}.

\noindent
\\
For DEER \citep{yang2025dynamic}, we use the official implementation and scripts provided by the authors\footnote{\url{https://github.com/iie-ycx/DEER}}. To ensure a fair and consistent comparison, we use the provided scripts and set think ratio  to 0.9 and max generated tokens to 16384 to align with our training configurations.

\noindent
\\
For NoThink \citep{nothinking}, we followed the paper to bypass the explicit reasoning process through prompting using "Okay, I think I have finished thinking.".

\noindent
\\
For LCPO \citep{L1}, we use the official implementation provided by the authors\footnote{\url{https://github.com/cmu-l3/l1}} and follow their L1-Exact setup. Training is conducted using GRPO with a specific length constraint embedded in the prompt, instructing the model to "Think for $N$ tokens". The actor model is optimized with a learning rate of $1 \times 10^{-6}$ and a global training batch size of 64. To ensure stability during policy evolution, we set the KL penalty coefficient ($\beta$) to 0.001. During the rollout phase, the model generates 8 independent reasoning trajectories for each prompt with a sampling temperature of 0.6. The maximum response length is configured to 4,096 tokens, adhering to the default parameters provided in the official repository

\noindent
\\
For O1-Pruner, it has a KL penalty coefficient of 0.001 and a learning rate of $1 \times 10^{-6}$. Rollout number is fixed at 5 and maximum response length is set to 16384 align with our training configurations. 

\noindent
\\
For PEAR, we followed the official codebase provided by the authors\footnote{\url{https://github.com/iNLP-Lab/PEAR}}. We set the batch size to 32 and the learning rate to $1 \times 10^{-6}$. The KL penalty coefficient is set to 0.001. The maximum response length is set to 16384 to align with our training configurations.

\section{Experimental Details for the Motivation Section} \label{apx:motivation}

This appendix provides additional details on the analysis of step-wise entropy trajectories used to motivate our method.

Given a generated CoT, we segment it into sentence-level steps $\{C_t\}_{t=1}^{T}$ and compute an entropy value $H_t$ for each step. We get a list of step-wise entropy
\begin{equation*}
\mathbf{H} = [H_1, H_2, \dots, H_T]
\end{equation*}
Let $\mathcal{T}$ denote the list of the reasoning steps indices in ascending order, which corresponds to the entropy vector $\mathbf{H} = [H_1, H_2, \dots, H_T]$.
\begin{equation*}
\mathcal{T} = [1, 2, \dots, T]
\end{equation*}
Then we can compute the Spearman correlation coefficient, $\rho$, is defined as the Pearson correlation coefficient between the rank variables of $\mathcal{T}$ and $\mathbf{H}$:
\begin{equation}
\rho = \frac{\text{cov}(\text{rank}(\mathcal{T}), \text{rank}(\mathbf{H}))}{\sigma_{\text{rank}(\mathcal{T})} \sigma_{\text{rank}(\mathbf{H})}}
\label{eq:spearman_rank}
\end{equation}
Where:
\begin{itemize}
    \item $\text{rank}(\mathcal{T})$ and $\text{rank}(\mathbf{H})$ are the vectors containing the ranks of the original observations.
    \item $\text{cov}(\cdot)$ is the covariance of the rank variables.
    \item $\sigma$ represents the standard deviation of the rank variables.
\end{itemize}

The final result $\rho$ (Spearman's rank correlation coefficient) directly quantify the stability and speed of the model's logical convergence. Smaller $\rho$ indicates that as the model progresses through reasoning steps $\mathcal{T}$, the entropy $H_t$ consistently decreases. As shown in the Figure \ref{fig:two_column} (right), all three models show that smaller $\rho$ and better downward entropy trends often relates to fewer tokens generated.

\section{Additional Empirical Analysis}
\subsection{Analysis on Momentum} \label{apx:momentum}

\begin{table}[h]
\centering

\scriptsize
\setlength{\tabcolsep}{2pt}
\begin{tabular}{lccccccccl}
\toprule
\multirow{2}{*}{\textbf{$\gamma$}} & \multicolumn{2}{c}{\textbf{AMC23}} & \multicolumn{2}{c}{\textbf{AIME24}} & \multicolumn{2}{c}{\textbf{MATH500}} & \multicolumn{2}{c}{\textbf{GPQA-D}}&\multirow{2}{*}{\textbf{AES}$\uparrow$}\\
\cmidrule(lr){2-3} \cmidrule(lr){4-5} \cmidrule(lr){6-7} \cmidrule(lr){8-9}
& Acc $\uparrow$ & Len $\downarrow$ & Acc $\uparrow$ & Len $\downarrow$ & Acc $\uparrow$ & Len $\downarrow$ & Acc $\uparrow$ & Len $\downarrow$ \\
\midrule
0.5  & 85.0 & 2180.1 & 40.0 & 6381.6 & 90.4 & 1441.6 & 36.4 & 2684.5 &1.04 \\
0.8  & 87.5 & 2288.1 & 26.7 & 5622.7 & 89.0 & 1441.5 & 33.8 & 5622.7 &0.66\\
\textbf{0.9}  & \textbf{87.5} & \textbf{2411.6} & \textbf{56.7} & \textbf{4639.3} & \textbf{90.6} & \textbf{1479.2} & \textbf{37.4} & \textbf{2452.4} & \textbf{1.53}\\
0.95 & 82.5 & 3160.9 & 56.7 & 5289.3 & 89.2 & 1445.7 & 35.4 & 3052.2 &1.29 \\
0.99 & 80.0 & 2371.0 & 46.7 & 5070.0 & 88.4 & 1489.8 & 38.9 & 2621.4 &1.12\\
\bottomrule
\end{tabular}
\caption{Analysis on the impact of the momentum $\gamma$ in Equation \ref{eq:momentum}. 
}
\label{tab:gamma_ablation}
\end{table}

To study the sensitivity to the momentum coefficient $\gamma$, we conduct ablation experiments with $\gamma \in \{0.2, 0.5, 0.8, 0.9, 0.95, 0.99\}$. Results are reported in Table~\ref{tab:gamma_ablation}.

Overall, model performance is highly sensitive to the choice of $\gamma$. A typical momentum value $\gamma = 0.9$ consistently achieves the best accuracy across challenging benchmarks, indicated by the highest AES score of 1.528. Smaller values ($\gamma \le 0.8$) lead to degraded reasoning accuracy, particularly on harder tasks such as AIME24. In contrast, overly large momentum ($\gamma \ge 0.95$) results in reduced accuracy and inflated generation length, indicating over-smoothing of step-wise entropy signals. These results suggest that $\gamma = 0.9$ offers the best balance between temporal stability and responsiveness.

\subsection{The Influence of LoRA Rank}
\begin{table}[ht]
    \centering
    \begin{minipage}{0.48\textwidth}
        \centering
        \resizebox{\textwidth}{!}{
        \begin{tabular}{lcccccccccc}
            \toprule
             & \multicolumn{2}{c}{AMC23} & \multicolumn{2}{c}{AIME24} & \multicolumn{2}{c}{MATH500} & \multicolumn{2}{c}{GPQA-D} & \multicolumn{2}{c}{Overall (Avg)} \\
            & Acc$\uparrow$ & Len$\downarrow$ & Acc$\uparrow$ & Len$\downarrow$ & Acc$\uparrow$ & Len$\downarrow$ & Acc$\uparrow$ & Len$\downarrow$ & Acc$\uparrow$ & Len$\downarrow$ \\
            \midrule
            Base & 80.00\% & 6597.63 & 43.33\% & 11751.53 & 85.00\% & 4210.51 & 24.20\% & 11336.66 & 58.13\% & 8474.08 \\
            \midrule
            Rank 16 & 85.00\% & 3071.23 & 46.67\% & 6970.80 & 90.00\% & 1961.92 & 38.38\% & 4718.47 & 65.01\% & 4180.61 \\
            Rank 32 & 87.50\% & 2411.63 & 56.67\% & 4639.30 & 90.60\% & 1479.16 & 37.37\% & 2452.43 & 68.04\% & 2745.63 \\
            Rank 64 & 87.50\% & 2130.05 & 53.33\% & 4256.60 & 90.80\% & 1381.37 & 39.39\% & 2174.15 & 67.76\% & 2485.54 \\
            \midrule
            Full FT & 87.50\% & 2863.15 & 53.33\% & 5208.20 & 91.20\% & 1739.81 & 39.90\% & 2267.26 & 67.98\% & 3019.61 \\
            \bottomrule
        \end{tabular}
        }
    \caption{Analysis on the impact of LoRA Rank $r$ on model performance.}\label{tab:lora_rank}
    \end{minipage}
    \hfill
\end{table}

Due to computational constraints, we performed training using LoRA. Here, we study the impact of the LoRA rank on our approach. 
Table~\ref{tab:lora_rank} summarizes the results. Several observations are noteworthy. First, performance improves with increasing LoRA rank but quickly saturates: ranks 32 and 64 achieve accuracies of 68.04\% and 67.76\%, respectively, comparable to full-parameter fine-tuning (67.98\%). Second, token length decreases as rank increases, indicating improved reasoning efficiency. Nevertheless, even Rank 16 substantially reduces the average length from 8474 to 4180 compared to the base model, suggesting that ETR’s effect is not limited to high-capacity adaptations. Third, LoRA performs on par with or better than full fine-tuning in efficiency. In particular, Rank 32 achieves a shorter average reasoning length (2745) than full fine-tuning (3019) while maintaining comparable accuracy. In practice, full fine-tuning requires more careful optimization and tends to overfit quickly, especially on smaller datasets, leading to unstable entropy dynamics and degraded generalization. To mitigate this issue, we apply early stopping in our experiments.

\begin{table}[ht]
\centering
\begin{minipage}{0.48\textwidth}
        \centering
        \resizebox{\textwidth}{!}{
\begin{tabular}{lcccccccccc}
\toprule
 & \multicolumn{2}{c}{\textbf{amc23}} & \multicolumn{2}{c}{\textbf{aime24}} & \multicolumn{2}{c}{\textbf{MATH500}} & \multicolumn{2}{c}{\textbf{GPQA-D}} & \multicolumn{2}{c}{\textbf{Overall}} \\
& Acc$\uparrow$ & Len$\downarrow$ & Acc$\uparrow$ & Len$\downarrow$ & Acc$\uparrow$ & Len$\downarrow$ & Acc$\uparrow$ & Len$\downarrow$ & Acc$\uparrow$ & Len$\downarrow$ \\
\midrule

\multicolumn{11}{c}{\textbf{DeepSeek-R1-Distill-Qwen-1.5B}} \\
\midrule
Untrained & 47.50\% & 9414.55 & 13.33\% & 13671.93 & 68.60\% & 6433.81 & 6.06\% & 14073.11 & 33.87\% & 10898.35 \\
ETR (FT) & \textbf{62.50\%} & \textbf{4481.43} & \textbf{13.33\%} & \textbf{10435.63} & \textbf{79.00\%} & \textbf{2220.44} & \textbf{18.69\%} & \textbf{7113.74} & \textbf{43.38\%} & \textbf{6062.81} \\
\midrule

\multicolumn{11}{c}{\textbf{DeepSeek-R1-Distill-14B}} \\
\midrule
Untrained & 87.50\% & 5926.85 & 33.33\% & 12623.07 & 87.40\% & 4000.64 & 41.92\% & 8878.32 & 62.54\% & 7857.22 \\
ETR (FT) & \textbf{92.50\%} & \textbf{2932.38} & \textbf{53.33\%} & \textbf{6087.90} & \textbf{90.00\%} & \textbf{1473.37} & \textbf{51.01\%} & \textbf{3210.10} & \textbf{71.71\%} & \textbf{3425.94} \\
\bottomrule
\end{tabular}
}
\caption{Model scale analysis on DeepSeek-R1-Distill-14B and DeepSeek-R1-Distill-Qwen-1.5B, showing consistent improvements in accuracy and reasoning efficiency using ETR, with substantial reductions in reasoning length across model sizes.}
\label{tab:model_scale}
\end{minipage}
\end{table}

\subsection{Model Scale Analysis}

To evaluate the scalability and generalizability of our approach, we conduct experiments on both a larger model, DeepSeek-R1-Distill-14B, and a smaller model, DeepSeek-R1-Distill-Qwen-1.5B. The results are summarized in Table~\ref{tab:model_scale}. 

On the larger 14B model, the proposed ETR strategy scales effectively, achieving consistent improvements in both accuracy and reasoning efficiency. Notably, ETR significantly reduces reasoning length across all benchmarks while improving overall accuracy, demonstrating strong scalability to higher-capacity models.

On the smaller 1.5B model, full fine-tuning substantially improves overall accuracy from 33.87\% to 43.38\% while reducing the average reasoning length from 10898 to 6063 tokens. In particular, on the MATH500 benchmark, the reasoning length is reduced by approximately 65\% (from 6433 to 2220 tokens), accompanied by a performance gain of 10.4\%. These results indicate that our approach remains effective even for smaller models, improving both reasoning quality and efficiency.

\subsection{Step-Level Entropy After ETR}
To better ilustrate how ETR effectively mitigates overthinking, we provide a detailed comparison of entropy dynamics on the same sample.

\begin{figure}[H]
    \centering
    \includegraphics[width=1\linewidth]{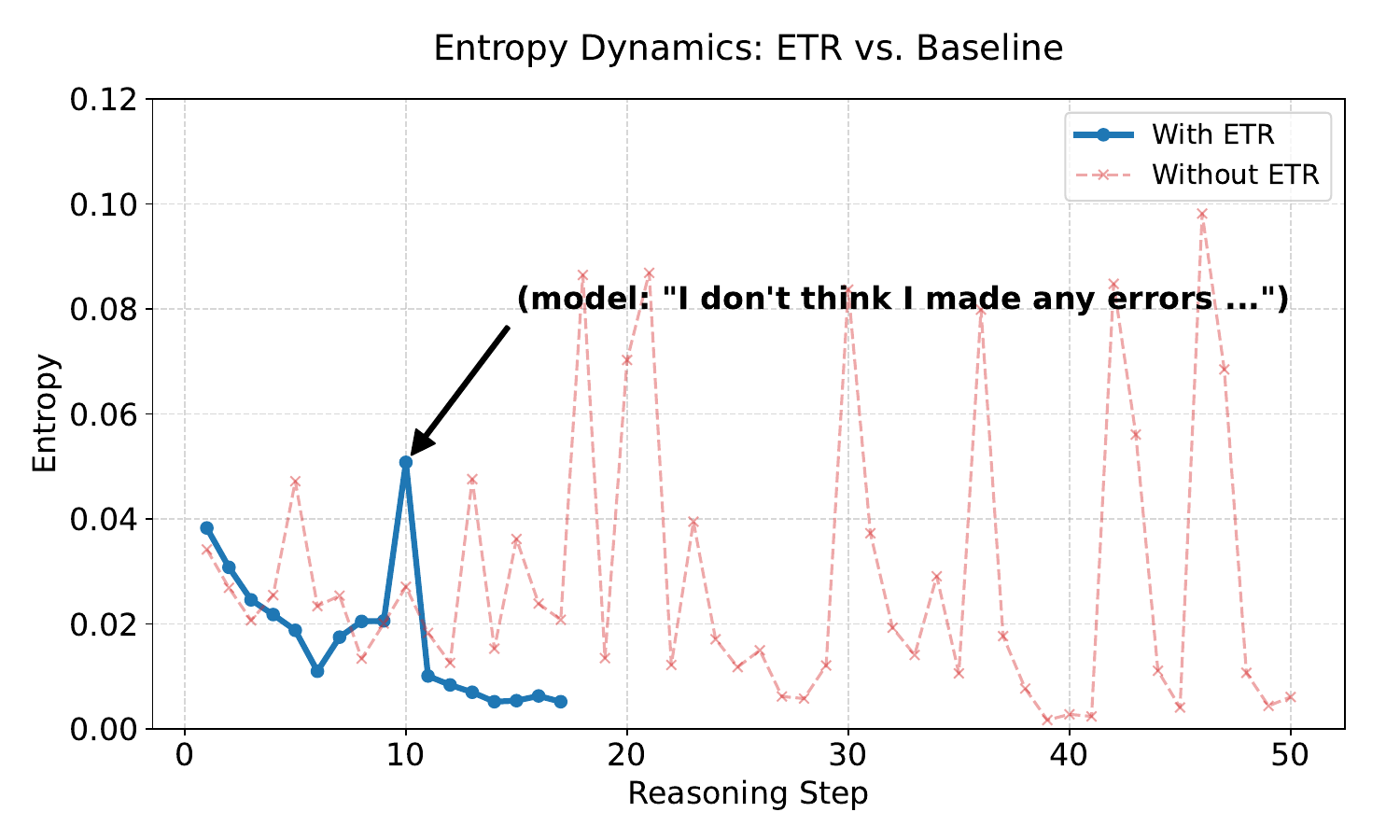}
    \vspace{-2em}
    \caption{Entropy trajectories during reasoning. ETR-trained models demonstrate efficient convergence and necessary self-reflection, whereas baseline models exhibit frequent entropy spikes and redundant reasoning steps.}
    \label{fig:entropy_correlation2}
\end{figure}

As shown in Figure~\ref{fig:entropy_correlation2}, ETR ensures that the reasoning process remains focused and leads to a definitive conclusion. Unlike baseline models that often experience frequent "entropy spikes", indicating excessive self-reflection and self-doubt, ETR ensures necessary reflection occurs efficiently, allowing the model to converge quickly to the final answer.

\subsection{Alternative Segmentation Strategies}
In this work, we segmented model's reasoning trajectory by "\textbackslash n\textbackslash n" and calculated average entropy based on it. Models typically generate this separator when completing a partial reasoning step. Empirically, these segments correspond to semantically coherent sub-steps in multi-step reasoning. 

To gain a more precise understanding of the ETR's impact on the model's internal dynamics, we extend our analysis to the token level segmentation. The results are shown below:

\begin{table}[ht]
    \centering
    \begin{minipage}{0.48\textwidth}
        \centering
        \resizebox{\textwidth}{!}{
        \begin{tabular}{lcccccccccc}
            \toprule
             & \multicolumn{2}{c}{AMC23} & \multicolumn{2}{c}{AIME24} & \multicolumn{2}{c}{MATH500} & \multicolumn{2}{c}{GPQA-D} & \multicolumn{2}{c}{Overall (Avg)} \\
            & Acc$\uparrow$ & Len$\downarrow$ & Acc$\uparrow$ & Len$\downarrow$ & Acc$\uparrow$ & Len$\downarrow$ & Acc$\uparrow$ & Len$\downarrow$ & Acc$\uparrow$ & Len$\downarrow$ \\
            \midrule
            Base & 80.00\% & 6597.63 & 43.33\% & 11751.53 & 85.00\% & 4210.51 & 24.20\% & 11336.66 & 58.13\% & 8474.08 \\
            Token\_Seg & 72.50\% & 1210.25 & 36.67\% & 3098.17 & 82.20\% & 916.04 & 33.84\% & 919.84 & 56.31\% & 1536.08 \\
            \midrule
            \textbf{ETR} & \textbf{87.50\%} & \textbf{2411.63} & \textbf{56.67\%} & \textbf{4639.30} & \textbf{90.60\%} & \textbf{1479.16} & \textbf{37.37\%} & \textbf{2452.43} & \textbf{68.04\%} & \textbf{2745.63} \\
            \bottomrule
        \end{tabular}
        }
        \caption{Performance comparison focusing on token-level segmentation vs. ETR.}
        \label{tab:token_level_comparison}
    \end{minipage}
\end{table}
While token-level regulation dramatically reduces length (8474 -> 1536 avg tokens), it also degrades accuracy (58.13\% -> 56.31\%). In contrast, ETR with step-level segmentation improves both accuracy (68.04\%) and efficiency (2745 tokens). 

The reason is that token-level segmentation is overly fine-grained. Individual tokens do not represent meaningful reasoning units, so entropy fluctuations mainly reflect lexical variability rather than logical uncertainty. This introduces noise and leads to over-compression, suppressing necessary inferential steps.

\subsection{Broadening Task Scopes: Beyond Mathematical Reasoning}
To evaluate generalization beyond math-related tasks, we tested ETR on the HumanEval coding benchmark\citep{chen2021codex}, which represents a substantially different reasoning domain. Importantly, our models were trained purely on mathematical reasoning data, making coding an out-of-distribution (OOD) setting.

\begin{table}[ht]
    \centering
    \begin{minipage}{0.48\textwidth}
        \centering
        \resizebox{\textwidth}{!}{
        \begin{tabular}{llcc}
            \toprule
            \textbf{Model} & \textbf{Method} & \textbf{Accuracy} $\uparrow$ & \textbf{Avg Length} $\downarrow$ \\
            \midrule
            Qwen3-4B & Base & 41.46\% & 3397.20 \\
            Qwen3-4B & ETR & \textbf{53.66\%} & \textbf{1838.60} \\
            \midrule
            DeepSeek-R1-Distill-7B & Base & 75.00\% & 3975.97 \\
            DeepSeek-R1-Distill-7B & ETR & \textbf{75.00\%} & \textbf{1057.99} \\
            \bottomrule
        \end{tabular}
        }
        \caption{Model Performance on HumanEval Coding Benchmark}
        \label{tab:model_comparison}
    \end{minipage}
\end{table}

Across both Qwen3-4B and DeepSeek-R1-Distill-7B, ETR consistently reduces reasoning length while maintaining or improving accuracy. For example, on Qwen3-4B, ETR improves accuracy from 41.46\% to 53.66\% while reducing average length from 3397 to 1838 tokens. On DeepSeek-R1-Distill-7B, ETR maintains accuracy (75.00\%) while reducing length from 3976 to 1058 tokens.

These results demonstrate that ETR transfers effectively to coding despite the absence of coding-specific RL training. This suggests that entropy trend regulation targets general reasoning convergence behavior rather than domain-specific patterns.

\subsection{The Influence of Entropy Trend Reward on Final Answer.}

In this section, we evaluate the quality and conciseness of the final response produced after the reasoning process. Unlike the internal reasoning trajectory, this segment represents the actual output presented to the user. While traditional reasoning models often carry over the verbosity of their internal chain-of-thought into the final output, our method helps to reduce such problem by presenting a final result that has higher information density and is free from ambiguity.
\subsubsection{Example 1}
\newtcolorbox{thoughtbox4}{colback=white!5, colframe=white!60!black, title=Final Answer Part , breakable}
\newcommand{\dashline}{%
    \noindent\begin{tikzpicture}
        \draw[dash pattern=on 3pt off 3pt, line width=0.5pt, blue!40!black] (0,0) -- (\linewidth,0);
    \end{tikzpicture}%
}
\begin{thoughtbox4}
\textbf{Before(408 Tokens):}
\\
To determine the remainder of $ 54 \mod 6 $, we apply the concept of modular arithmetic, which involves finding the remainder when one number is divided by another.
\\
---\\
\#\#\# Step 1: Understand the Problem
\\\\
We are asked to compute the remainder when 54 is divided by 6. In mathematical terms, this is:
\\\\
$54 \mod 6$
\\\\
This means we want to find the integer $ r $ such that:
\\\\
$54 = 6q + r \quad \text{where } 0 \le r < 6$
\\\\
Here, $ q $ is the quotient, and $ r $ is the remainder.
\\\\
---
\\\\
\#\#\# Step 2: Perform the Division
\\\\
We divide 54 by 6:\\
$54 \div 6 = 9$
This division is exact, meaning there is no remainder. Therefore, the remainder $ r $ is:
\\\\
$r = 54 - (6 \times 9) = 54 - 54 = 0$
\\\\---\\\\\#\# Step 3: Confirm with Alternative Methods\\\\- **Using the Division Algorithm**: Since $ 54 = 6 \times 9 + 0 $, the remainder is clearly 0.
\\
- **Using Properties of Divisibility**: 54 is divisible by both 2 and 3 (since it's even and the sum of its digits is 9, which is divisible by 3), so it is divisible by 6.
\\
- **Breaking Down the Number**: $ 54 = 50 + 4 $, and $ 50 \mod 6 = 2 $, $ 4 \mod 6 = 4 $, so $ (2 + 4) \mod 6 = 6 \mod 6 = 0 $.
\\\\
All these methods confirm the same result.
\\\\
---
\\\\
\#\#\# Final Answer
\\
$$\boxed{0}$$
\dashline

\textbf{After(64 Tokens)}\\
To determine the remainder of $ 54 \mod 6 $, we perform the division of 54 by 6. Since $ 6 \times 9 = 54 $, the division is exact with no remainder. Therefore, the remainder is:
\\
$$\boxed{0}$$
\end{thoughtbox4}

\subsubsection{Example 2}
\begin{thoughtbox4}
\textbf{Before(386 Tokens):}\\
To find the **dot product** of two vectors, we use the formula:

$$\mathbf{u} \cdot \mathbf{v} = u_1 v_1 + u_2 v_2 + u_3 v_3$$
\\
where $\mathbf{u} = 
\begin{pmatrix} 
u_1 \\\\ u_2 \\\\ u_3 \end{pmatrix}$ and $\mathbf{v} = \begin{pmatrix} v_1 \\\\ v_2 \\\\ v_3 \end{pmatrix}$ are two vectors in $\mathbb{R}^3$.
\\\\
---
\\\\
\#\#\# Step 1: Identify the components of the vectors
\\\\
Given:
\\\\
$$\mathbf{u} = \begin{pmatrix} -2 \\\\ 0 \\\\ 7 \end{pmatrix}, \quad \mathbf{v} = \begin{pmatrix} 3 \\\\ 4 \\\\ -5 \end{pmatrix}$$
\\\\
So, the components are:\\\\
- $u_1 = -2$, $u_2 = 0$, $u_3 = 7$\\
- $v_1 = 3$, $v_2 = 4$, $v_3 = -5$
\\\\---\\\\
\#\#\# Step 2: Compute the dot product\\\\$$\mathbf{u} \cdot \mathbf{v} = (-2)(3) + (0)(4) + (7)(-5)$$
\\\\Now compute each term:\\\\- $(-2)(3) = -6$\\- $(0)(4) = 0$\\- $(7)(-5) = -35$\\\\Add them together:\\$$-6 + 0 + (-35) = -41$$---\\\\\#\#\# Final Answer\\$$\boxed{-41}$$
\dashline

\textbf{After(127 Tokens):}\\
To find the dot product of the vectors  \\$$\begin{pmatrix} -2 \\\\ 0 \\\\ 7 \end{pmatrix} \quad \text{and} \quad \begin{pmatrix} 3 \\\\ 4 \\\\ -5 \end{pmatrix},\\$$  multiply the corresponding components and then sum the results:\\\\
\begin{equation*}
\begin{aligned}
(-2)(3) + (0)(4) + (7)(-5) &= -6 + 0 - 35 \\
&= -41
\end{aligned}
\end{equation*}
Thus, the dot product is \\$$\boxed{-41}$$
\end{thoughtbox4}
\section{Additional Responses of ETR Models}

\subsection{Qwen3-8B}
\begin{promptbox}
A regular octagon has the same perimeter as the regular hexagon shown here with side length 16 cm.  How long is each side of the octagon?
\end{promptbox}
\begin{errorbox}
(Original Model Output Tokens: \textbf{5140})
\end{errorbox}
\begin{thoughtbox}
\small 
<think>\\
Okay, so I need to find the side length of a regular octagon that has the same perimeter as a regular hexagon with side length 16 cm. Let me start by recalling some basics about regular polygons. 
\\ \\
First, a regular hexagon has all sides equal and all internal angles equal. Similarly, a regular octagon also has all sides equal. The perimeter of a regular polygon is just the number of sides multiplied by the length of one side. 
\\ \\
So, the perimeter of the hexagon would be 6 times the side length. Since the side length is given as 16 cm, the perimeter of the hexagon is 6 * 16 cm. Let me calculate that: 6 * 16 is 96 cm. So the perimeter is 96 cm.
\\ \\
Now, the regular octagon has the same perimeter. Let me denote the side length of the octagon as 's'. Since an octagon has 8 sides, its perimeter would be 8 * s. Therefore, setting this equal to the perimeter of the hexagon: 8s = 96 cm. To find 's', I can divide both sides by 8. 
\\ \\
Let me do that: 96 divided by 8. Hmm, 8*12 is 96, so s = 12 cm. Wait, is that right? Let me check again. 8 times 12 is 96, yes. So each side of the octagon would be 12 cm. 
\\ \\
But wait, let me make sure I didn't mix up the number of sides. Hexagon has 6 sides, octagon has 8. So yes, if the perimeters are equal, then 6*16 = 8*s. So s = (6*16)/8. Which is (96)/8 = 12. Yep, that seems correct. 
\\ \\
I think that's all there is to it. The Asymptote code provided probably just draws the hexagon with a label on one of its sides, but since the problem states the hexagon has side length 16 cm, I don't need to worry about the diagram specifics. 
\\ \\
Therefore, the answer should be 12 cm.
\\ \\
**Final Answer**
The length of each side of the octagon is \boxed{12} centimeters.
\\</think>\\

To determine the side length of a regular octagon that has the same perimeter as a regular hexagon with side length 16 cm, we start by calculating the perimeter of the hexagon.

A regular hexagon has 6 sides, so its perimeter is:
$$
6 \times 16 = 96 \text{ cm}
$$

A regular octagon has 8 sides. Let the side length of the octagon be $ s $. Its perimeter is:
$$
8s
$$

Since the perimeters are equal, we set:
$$
8s = 96
$$

Solving for $ s $:
$$
s = \frac{96}{8} = 12
$$

Thus, the length of each side of the octagon is $\boxed{12}$ centimeters.
\end{thoughtbox}

\subsection{Qwen3-4B}

\begin{promptbox}
What is the least positive integer multiple of 30 that can be written with only the digits 0 and 2?
\end{promptbox}

\begin{errorbox}
(Original Model Output Tokens: \textbf{9170})
\end{errorbox}

\begin{thoughtbox2}
\small 
<think>\\
Okay, so I need to find the smallest positive integer multiple of 30 that only has digits 0 and 2. Let me think. First, since it's a multiple of 30, it has to be divisible by both 3 and 10. Divisible by 10 means the number has to end with a 0. So the last digit is 0. Then, for divisibility by 3, the sum of the digits has to be a multiple of 3. Since the digits can only be 0 or 2, the sum will be 2 times the number of 2s in the number. So the number of 2s has to be a multiple of 3. Therefore, the number must have 3, 6, 9, etc., number of 2s. But since we want the smallest such number, probably 3 twos. But also, the number has to end with 0. So maybe something like 2220? Let me check. 2220 divided by 30 is 74, which is an integer. Wait, is 2220 a multiple of 30? Let me check. 2220 divided by 10 is 222, which is divisible by 3 because 2+2+2=6, which is divisible by 3. So yes, 2220 is a multiple of 30. But is there a smaller number? Let me think. The number has to end with 0. So possible numbers are 20, 200, 220, 2000, 2200, 2020, 2220, etc. Let's check each of these. 20: sum of digits is 2, not divisible by 3. 200: sum is 2, same. 220: sum is 2+2=4, not divisible by 3. 2000: sum is 2, no. 2200: sum is 2+2=4, no. 2020: sum is 2+2=4, no. 2220: sum is 2+2+2=6, which is divisible by 3. So 2220 is the first one. Wait, but is there a smaller number? Like, maybe a number with fewer digits? 2220 is four digits. Is there a three-digit number? The three-digit numbers ending with 0 are 200, 220, 200. But their digit sums are 2, 4, 2. None are divisible by 3. So no. How about a four-digit number? 2220 is the first one. So maybe 2220 is the answer. But let me check if there's a smaller four-digit number. For example, 2020, 2002, but those don't end with 0. Wait, the number has to end with 0. So the four-digit numbers ending with 0 are 2000, 2020, 2200, 2220. We checked those. 2000: sum 2. 2020: sum 4. 2200: sum 4. 2220: sum 6. So 2220 is the first. Therefore, the answer is 2220. Wait, but let me check if there's a number with more digits but smaller in value. Wait, no, because 2220 is four digits. Any number with more digits would be larger. So I think 2220 is the answer. But let me confirm by dividing 2220 by 30. 30 times 74 is 2220. Yes. So I think that's it.
\\ \\ 
**Final Answer**
The least positive integer multiple of 30 that can be written with only the digits 0 and 2 is $\boxed{2220}$.
\\
</think>
\\
To find the least positive integer multiple of 30 that can be written using only the digits 0 and 2, we must consider the divisibility rules for 30. A number is divisible by 30 if it is divisible by both 3 and 10.
\\ \\
- **Divisibility by 10**: The number must end in 0.
- **Divisibility by 3**: The sum of the digits must be a multiple of 3. Since the only digits allowed are 0 and 2, the sum of the digits will be $2 \times (\text{number of 2s})$. Therefore, the number of 2s must be a multiple of 3.
\\ \\ 
The smallest such number is formed by using three 2s and one 0 at the end, resulting in the number **2220**. Checking:
- It ends in 0, so it is divisible by 10.
- The sum of its digits is $2 + 2 + 2 + 0 = 6$, which is divisible by 3.
\\ \\ 
Thus, 2220 is the smallest number meeting the criteria.

$$
\boxed{2220}
$$
\end{thoughtbox2}

\subsection{DeepSeek-R1-Distill-Qwen-7B}

\begin{promptbox}
What is the concentration of calcium ions in a solution containing 0.02 M stochiometric Ca-EDTA complex (we assume that the pH is ideal, T = 25 °C). KCa-EDTA = $5x10^10$.
\\ \\
A. $6.3x10^-7 M$ \\
B. $1.0x10^-2 M$ \\
C. $5.0x10^-3 M$ \\
D. $2.0x10^-2 M$ \\
\end{promptbox}

\begin{errorbox}
(Original Model Output Tokens: \textbf{16384})
\end{errorbox}

\begin{thoughtbox3}
\small 
</think>\\
Okay, so I have this problem here about the concentration of calcium ions in a solution. Let me try to figure it out step by step. 
\\ \\
First, the problem says there's a 0.02 M stochiometric Ca-EDTA complex. I think that means the concentration of the complex is 0.02 M. They also mention that the pH is ideal and the temperature is 25°C. The formation constant, Kf, is given as $5 \times 10^{10}$.
\\ \\
Hmm, formation constant, Kf, is the equilibrium constant for the formation of the complex between calcium ions and EDTA. The reaction should be something like \ce{Ca^2+ + EDTA^4- <=> [Ca-EDTA]^2-} complex. 
\\ \\
So, the formation constant Kf is equal to $[Ca-EDTA]/([Ca^2+][EDTA^{4-}])$. Since it's a stochiometric complex, I assume the concentration of $Ca^{2+}$ and $EDTA^{4-}$ are equal because they react in a 1:1 ratio. Let me denote the concentration of $Ca^{2+}$ as 'c'. Then, the concentration of $EDTA^{4-}$ is also 'c'.
\\ \\
The total concentration of the complex is 0.02 M, which is equal to [Ca-EDTA]. So, $Kf = 0.02 / (c * c) = 0.02 / c^{2}$.
\\ \\
Given that Kf is $5 \times 10^{10}$, I can set up the equation: $5 \times 10^{10} = 0.02 / c^2$.
\\ \\
To solve for $c^2$, I rearrange the equation: $c^2 = 0.02 / (5 \times 10^{10})$. Let me calculate that. 0.02 divided by 5 is 0.004, and then divided by $10^{10}$ is $4x10^{-13}$. So, $c^2 = 4 \times 10^{-13}$.
\\ \\
Taking the square root of both sides, 
$c = \sqrt{4 \times 10^{-13}} = 2 \times 10^{-6.5}$. 
Wait, $10^{-6.5}$ is the same as $10^{-6} \times 10^{-0.5}$, 
which is approximately $3.16 \times 10^{-7}$. 
So, $2 \times 3.16 \times 10^{-7}$ is about $6.32 \times 10^{-7}$.
\\ \\
Looking at the options, A is $6.3 \times 10^{-7} M$, which matches my calculation. So, I think the answer is A.
\\ \\
**Final Answer** \\
\boxed{A} 
\\
</think>
The problem involves finding the concentration of calcium ions in a solution containing a 0.02 M stochiometric Ca-EDTA complex. The formation constant \( K_f \) is given as \( 5 \times 10^{10} \).
\\ \\
1. The reaction is \( \text{Ca}^{2+} + \text{EDTA}^{4-} \rightleftharpoons \text{Ca-EDTA complex} \). \\
2. The formation constant \( K_f \) is given by \( K_f = \frac{[\text{Ca-EDTA}]}{[\text{Ca}^{2+}][\text{EDTA}^{4-}]} \). \\
3. Since the complex is stochiometric, the concentrations of \( \text{Ca}^{2+} \) and \( \text{EDTA}^{4-} \) are equal, denoted as \( c \). \\
4. The total concentration of the complex is 0.02 M, so \( K_f = \frac{0.02}{c^2} \). \\
5. Solving for \( c^2 \): \( c^2 = \frac{0.02}{5 \times 10^{10}} = 4 \times 10^{-13} \). \\
6. Taking the square root: \( c = \sqrt{4 \times 10^{-13}} = 2 \times 10^{-6.5} \approx 6.3 \times 10^{-7} \).
\\ \\
Thus, the concentration of calcium ions is \( 6.3 \times 10^{-7} \) M.
\[
\boxed{A}
\]
\end{thoughtbox3}

%% file: custom.bib
@article{prompt1,
  title={Chain of draft: Thinking faster by writing less},
  author={Xu, Silei and Xie, Wenhao and Zhao, Lingxiao and He, Pengcheng},
  journal={arXiv preprint arXiv:2502.18600},
  year={2025}
}

@article{prompt2,
  title={How well do llms compress their own chain-of-thought? a token complexity approach},
  author={Lee, Ayeong and Che, Ethan and Peng, Tianyi},
  journal={arXiv preprint arXiv:2503.01141},
  year={2025}
}

@article{sft1,
  title={Tokenskip: Controllable chain-of-thought compression in llms},
  author={Xia, Heming and Leong, Chak Tou and Wang, Wenjie and Li, Yongqi and Li, Wenjie},
  journal={arXiv preprint arXiv:2502.12067},
  year={2025}
}

@article{sft2,
  title={Limo: Less is more for reasoning},
  author={Ye, Yixin and Huang, Zhen and Xiao, Yang and Chern, Ethan and Xia, Shijie and Liu, Pengfei},
  journal={arXiv preprint arXiv:2502.03387},
  year={2025}
}

@article{sft3,
  title={Can language models learn to skip steps?},
  author={Liu, Tengxiao and Guo, Qipeng and Hu, Xiangkun and Jiayang, Cheng and Zhang, Yue and Qiu, Xipeng and Zhang, Zheng},
  journal={Advances in Neural Information Processing Systems},
  volume={37},
  pages={45359--45385},
  year={2024}
}

@article{survey1,
  title={Don't Overthink It: A Survey of Efficient R1-style Large Reasoning Models},
  author={Yue, Linan and Du, Yichao and Wang, Yizhi and Gao, Weibo and Yao, Fangzhou and Wang, Li and Liu, Ye and Xu, Ziyu and Liu, Qi and Di, Shimin and others},
  journal={arXiv preprint arXiv:2508.02120},
  year={2025}
}

@article{survey2,
  title={A survey of efficient reasoning for large reasoning models: Language, multimodality, and beyond},
  author={Qu, Xiaoye and Li, Yafu and Su, Zhaochen and Sun, Weigao and Yan, Jianhao and Liu, Dongrui and Cui, Ganqu and Liu, Daizong and Liang, Shuxian and He, Junxian and others},
  journal={arXiv preprint arXiv:2503.21614},
  year={2025}
}

@article{survey3,
  title={Stop overthinking: A survey on efficient reasoning for large language models},
  author={Sui, Yang and Chuang, Yu-Neng and Wang, Guanchu and Zhang, Jiamu and Zhang, Tianyi and Yuan, Jiayi and Liu, Hongyi and Wen, Andrew and Zhong, Shaochen and Zou, Na and others},
  journal={arXiv preprint arXiv:2503.16419},
  year={2025}
}

@article{nothinking,
  title={Reasoning models can be effective without thinking},
  author={Ma, Wenjie and He, Jingxuan and Snell, Charlie and Griggs, Tyler and Min, Sewon and Zaharia, Matei},
  journal={arXiv preprint arXiv:2504.09858},
  year={2025}
}

@article{step_entropy,
  title={Compressing chain-of-thought in llms via step entropy},
  author={Li, Zeju and Zhong, Jianyuan and Zheng, Ziyang and Wen, Xiangyu and Xu, Zhijian and Cheng, Yingying and Zhang, Fan and Xu, Qiang},
  journal={arXiv preprint arXiv:2508.03346},
  year={2025}
}

@article{cot,
  title={Chain-of-thought prompting elicits reasoning in large language models},
  author={Wei, Jason and Wang, Xuezhi and Schuurmans, Dale and Bosma, Maarten and Xia, Fei and Chi, Ed and Le, Quoc V and Zhou, Denny and others},
  journal={Advances in neural information processing systems},
  volume={35},
  pages={24824--24837},
  year={2022}
}

@article{O1-pruner,
  title={O1-pruner: Length-harmonizing fine-tuning for o1-like reasoning pruning},
  author={Luo, Haotian and Shen, Li and He, Haiying and Wang, Yibo and Liu, Shiwei and Li, Wei and Tan, Naiqiang and Cao, Xiaochun and Tao, Dacheng},
  journal={arXiv preprint arXiv:2501.12570},
  year={2025}
}

@article{L1,
  title={L1: Controlling how long a reasoning model thinks with reinforcement learning},
  author={Aggarwal, Pranjal and Welleck, Sean},
  journal={arXiv preprint arXiv:2503.04697},
  year={2025}
}

@article{deepmath,
  title={Deepmath-103k: A large-scale, challenging, decontaminated, and verifiable mathematical dataset for advancing reasoning},
  author={He, Zhiwei and Liang, Tian and Xu, Jiahao and Liu, Qiuzhi and Chen, Xingyu and Wang, Yue and Song, Linfeng and Yu, Dian and Liang, Zhenwen and Wang, Wenxuan and others},
  journal={arXiv preprint arXiv:2504.11456},
  year={2025}
}

@article{hendrycks2021math,
  title={Measuring mathematical problem solving with the math dataset, 2021},
  author={Hendrycks, Dan and Burns, Collin and Kadavath, Saurav and Arora, Akul and Basart, Steven and Tang, Eric and Song, Dawn and Steinhardt, Jacob},
  journal={URL https://arxiv. org/abs/2103.03874},
  volume={2},
  year={2024}
}

@inproceedings{rein2023gpqa,
  title={Gpqa: A graduate-level google-proof q\&a benchmark},
  author={Rein, David and Hou, Betty Li and Stickland, Asa Cooper and Petty, Jackson and Pang, Richard Yuanzhe and Dirani, Julien and Michael, Julian and Bowman, Samuel R},
  booktitle={First Conference on Language Modeling},
  year={2024}
}

@article{yang2025dynamic,
  title={Dynamic early exit in reasoning models},
  author={Yang, Chenxu and Si, Qingyi and Duan, Yongjie and Zhu, Zheliang and Zhu, Chenyu and Li, Qiaowei and Chen, Minghui and Lin, Zheng and Wang, Weiping},
  journal={arXiv preprint arXiv:2504.15895},
  year={2025}
}

@article{huang2025pear,
  title={PEAR: Phase Entropy Aware Reward for Efficient Reasoning},
  author={Huang, Chen and Lu, Wei and Zhang, Wenxuan},
  journal={arXiv preprint arXiv:2510.08026},
  year={2025}
}

@inproceedings{sheng2024hybridflow,
  title={Hybridflow: A flexible and efficient rlhf framework},
  author={Sheng, Guangming and Zhang, Chi and Ye, Zilingfeng and Wu, Xibin and Zhang, Wang and Zhang, Ru and Peng, Yanghua and Lin, Haibin and Wu, Chuan},
  booktitle={Proceedings of the Twentieth European Conference on Computer Systems},
  pages={1279--1297},
  year={2025}
}

@article{GRPO,
  title={Deepseekmath: Pushing the limits of mathematical reasoning in open language models},
  author={Shao, Zhihong and Wang, Peiyi and Zhu, Qihao and Xu, Runxin and Song, Junxiao and Bi, Xiao and Zhang, Haowei and Zhang, Mingchuan and Li, YK and Wu, Yang and others},
  journal={arXiv preprint arXiv:2402.03300},
  year={2024}
}

@article{ppo,
  title={Proximal policy optimization algorithms},
  author={Schulman, John and Wolski, Filip and Dhariwal, Prafulla and Radford, Alec and Klimov, Oleg},
  journal={arXiv preprint arXiv:1707.06347},
  year={2017}
}

@article{deepseekr12025incentivizing,
  title={Deepseek-r1: Incentivizing reasoning capability in llms via reinforcement learning},
  author={Guo, Daya and Yang, Dejian and Zhang, Haowei and Song, Junxiao and Zhang, Ruoyu and Xu, Runxin and Zhu, Qihao and Ma, Shirong and Wang, Peiyi and Bi, Xiao and others},
  journal={arXiv preprint arXiv:2501.12948},
  year={2025}
}

@article{openai2024o1systemcard,
  title={Openai o1 system card},
  author={Jaech, Aaron and Kalai, Adam and Lerer, Adam and Richardson, Adam and El-Kishky, Ahmed and Low, Aiden and Helyar, Alec and Madry, Aleksander and Beutel, Alex and Carney, Alex and others},
  journal={arXiv preprint arXiv:2412.16720},
  year={2024}
}

@article{shannon1948mathematical,
  title={A mathematical theory of communication},
  author={Shannon, Claude E},
  journal={The Bell system technical journal},
  volume={27},
  number={3},
  pages={379--423},
  year={1948},
  publisher={Nokia Bell Labs}
}

@article{huang2025efficient,
  title={Efficient reasoning for large reasoning language models via certainty-guided reflection suppression},
  author={Huang, Jiameng and Lin, Baijiong and Feng, Guhao and Chen, Jierun and He, Di and Hou, Lu},
  journal={arXiv preprint arXiv:2508.05337},
  year={2025}
}

@article{agarwal2025unreasonable,
  title={The unreasonable effectiveness of entropy minimization in llm reasoning},
  author={Agarwal, Shivam and Zhang, Zimin and Yuan, Lifan and Han, Jiawei and Peng, Hao},
  journal={arXiv preprint arXiv:2505.15134},
  year={2025}
}

@misc{aime24,
  title={American invitational mathematics examination - aime.},
  author={MAA.},
journal={In American Invitational Mathematics Examination - AIME.},
  year={2024}
}

@misc{amc23,
  author       = {MAA.},
  title        = {American mathematics competitions.},
  year         = {2023},
}

@inproceedings{hu2021lora,
  title={LoRA: Low-Rank Adaptation of Large Language Models},
  author={Hu, Edward J and Wallis, Phillip and Allen-Zhu, Zeyuan and Li, Yuanzhi and Wang, Shean and Wang, Lu and Chen, Weizhu and others},
  booktitle={International Conference on Learning Representations},
year={2022},
}

@article{zheng2024sglang,
  title={Sglang: Efficient execution of structured language model programs},
  author={Zheng, Lianmin and Yin, Liangsheng and Xie, Zhiqiang and Sun, Chuyue Livia and Huang, Jeff and Yu, Cody Hao and Cao, Shiyi and Kozyrakis, Christos and Stoica, Ion and Gonzalez, Joseph E and others},
  journal={Advances in neural information processing systems},
  volume={37},
  pages={62557--62583},
  year={2024}
}

@inproceedings{lightman2023let,
  title={Let's verify step by step},
  author={Lightman, Hunter and Kosaraju, Vineet and Burda, Yuri and Edwards, Harrison and Baker, Bowen and Lee, Teddy and Leike, Jan and Schulman, John and Sutskever, Ilya and Cobbe, Karl},
  booktitle={The Twelfth International Conference on Learning Representations},
  year={2023}
}

@article{chen2021codex,
  title={Evaluating Large Language Models Trained on Code},
  author={Mark Chen and Jerry Tworek and Heewoo Jun and Qiming Yuan and Henrique Ponde de Oliveira Pinto and Jared Kaplan and Harri Edwards and Yuri Burda and Nicholas Joseph and Greg Brockman and Alex Ray and Raul Puri and Gretchen Krueger and Michael Petrov and Heidy Khlaaf and Girish Sastry and Pamela Mishkin and Brooke Chan and Scott Gray and Nick Ryder and Mikhail Pavlov and Alethea Power and Lukasz Kaiser and Mohammad Bavarian and Clemens Winter and Philippe Tillet and Felipe Petroski Such and Dave Cummings and Matthias Plappert and Fotios Chantzis and Elizabeth Barnes and Ariel Herbert-Voss and William Hebgen Guss and Alex Nichol and Alex Paino and Nikolas Tezak and Jie Tang and Igor Babuschkin and Suchir Balaji and Shantanu Jain and William Saunders and Christopher Hesse and Andrew N. Carr and Jan Leike and Josh Achiam and Vedant Misra and Evan Morikawa and Alec Radford and Matthew Knight and Miles Brundage and Mira Murati and Katie Mayer and Peter Welinder and Bob McGrew and Dario Amodei and Sam McCandlish and Ilya Sutskever and Wojciech Zaremba},
  year={2021},
  eprint={2107.03374},
  archivePrefix={arXiv},
  primaryClass={cs.LG}
}
